\documentclass[sigconf,edbt]{acmart-edbt2024}
\usepackage{algorithm}
\usepackage{algpseudocode}
\usepackage{caption}
\usepackage{subcaption}
\usepackage{array}
\usepackage{multirow}
\usepackage{float}
\usepackage{boldline}
\algtext*{EndWhile}
\algtext*{EndIf}
\algtext*{EndFor}
\usepackage[normalem]{ulem}
\usepackage{booktabs} 

\AtBeginDocument{%
  \providecommand\BibTeX{{\rm B\kern-.05em{\sc i\kern-.025em b}\kern-.08em
    T\kern-.1667em\lower.7ex\hbox{E}\kern-.125emX}}}

\setcopyright{rightsretained}

\acmDOI{}

\acmConference[EDBT 2024]{27th International Conference on Extending Database Technology (EDBT)}{25th March-28th March, 2024}{Paestum, Italy}
\acmYear{2024}

\acmISBN{978-3-89318-095-0}

\begin{document}

\title{Spatial-temporal Forecasting for Regions without Observations}

\author{Xinyu Su}
\affiliation{%
  \institution{The University of Melbourne}
  \country{Australia}}
\email{suxs3@student.unimelb.edu.au}

\author{Jianzhong Qi}
\affiliation{%
  \institution{The University of Melbourne}
  \country{Australia}}
\email{jianzhong.qi@unimelb.edu.au}

\author{Egemen Tanin}
\affiliation{%
  \institution{The University of Melbourne}
  \country{Australia}}
\email{etanin@unimelb.edu.au}

\author{Yanchuan Chang}
\affiliation{%
  \institution{The University of Melbourne}
  \country{Australia}}
\email{yanchuan.chang@unimelb.edu.au}

\author{Majid Sarvi}
\affiliation{%
  \institution{The University of Melbourne}
  \country{Australia}}
\email{majid.sarvi@unimelb.edu.au}

\renewcommand{\shortauthors}{X. Su, et al.}

\newcommand{\model}{STSM}

\begin{abstract}
  Spatial-temporal forecasting plays an important role in many real-world applications, such as traffic forecasting, air pollutant forecasting, crowd-flow forecasting, and so on. State-of-the-art spatial-temporal forecasting models take data-driven approaches and rely heavily on data availability. Such models suffer from accuracy issues when data is incomplete, which is common in reality due to the heavy costs of deploying and maintaining sensors for data collection. A few recent studies attempted to address the issue of incomplete data. They typically assume some data availability in a region of interest either for a short period or at a few locations.
In this paper, we further study spatial-temporal forecasting for a region of interest without any historical observations, to address scenarios such as unbalanced region development, progressive deployment of sensors or lack of open data. We propose a model named \model\ for the task. The model takes a contrastive learning-based approach to learn spatial-temporal patterns from adjacent regions that have recorded data. Our key insight is to learn from the locations that resemble those in the region of interest, and we propose a selective masking strategy to enable the learning. As a result, our model outperforms adapted state-of-the-art models, reducing errors consistently over both traffic and air pollutant forecasting tasks. The source code is available at https://github.com/suzy0223/STSM.
\end{abstract}

\maketitle

\section{Introduction}
Spatial-temporal forecasting is an important component of many real-world applications, e.g., Intelligent Transportation Systems (ITS) and air quality monitoring systems. Current state-of-the-art methods for spatial-temporal forecasting are data-driven. They leverage sequential models, e.g., 1-D convolutional neural networks (CNN)~\cite{stsgcn,gamcn,graphwavenet} or recurrent neural networks (RNN)~\cite{dcrnn}, to capture temporal features, along with spatial models, e.g., graph neural networks (GNN), which model spatial relations~\cite{stsgcn,gamcn,graphwavenet}. However, \emph{data scarcity} is ubiquitous due to high deployment and maintenance costs of sensors and unstable transmission mediums. Hence, developing a model that enables accurate forecasting without complete historical data is critical. 

\begin{figure}[h!]
         \centering
  \includegraphics[width=\linewidth]{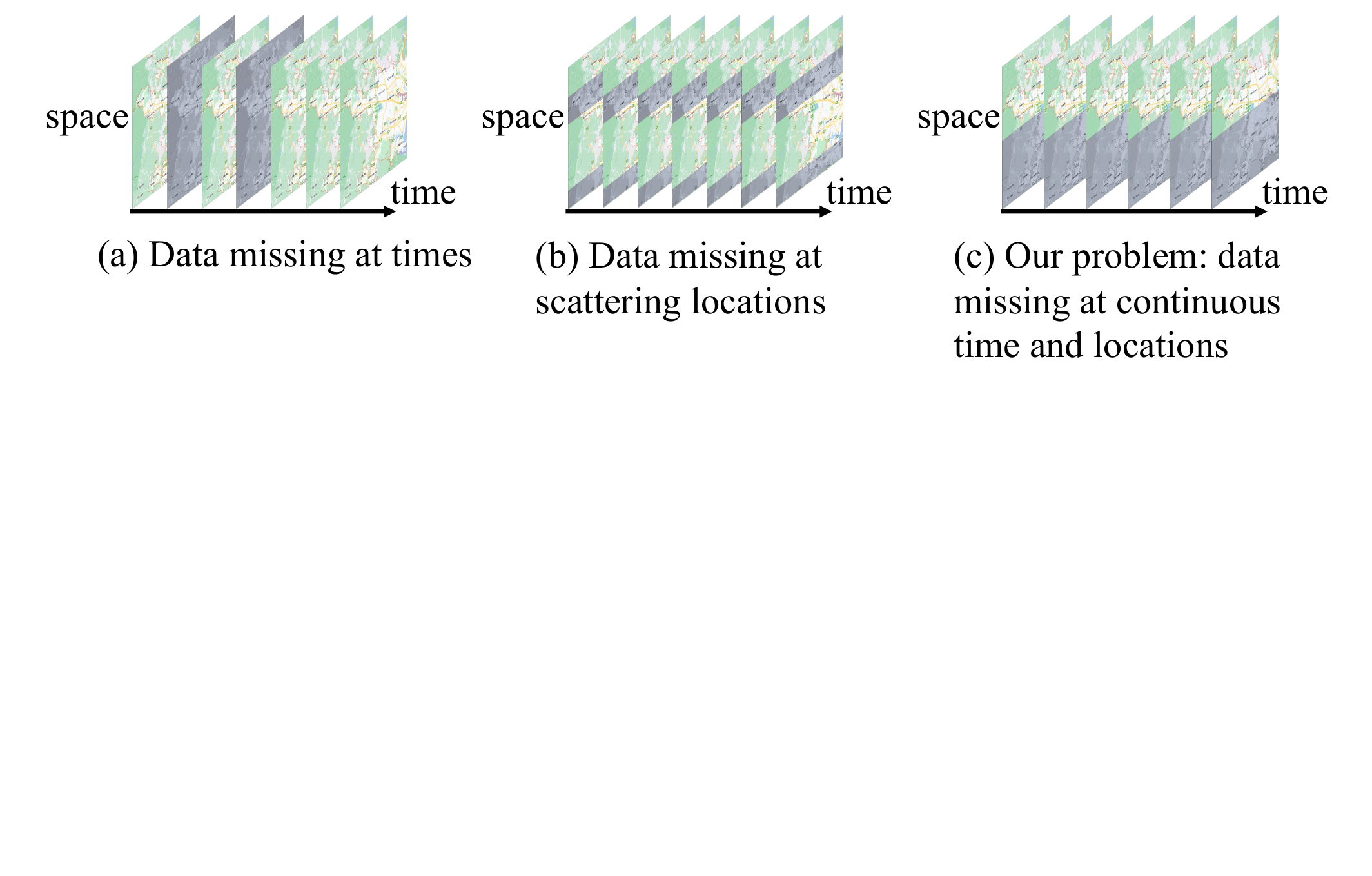}
  \vspace{-38mm}
         \caption{Problem setting comparison. Coloured maps and grey maps indicate data observed and unobserved, respectively. Our focus is Case (c).}
         \label{fig:problem}
\end{figure}

Prior attempts to address the data scarcity issue for spatial-temporal tasks fall into two categories: (1)~\emph{data missing at times}~\cite{mdgcn,ge-gan,sa-gain,dastnet,surrounding_imputation_clustering}: observations at locations of interest are incomplete for all time due to complex environments and/or faulty sensors, or a short sensor deployment time; (2)~\emph{data missing at (scattering) locations}~\cite{kcn,ignnk,increase}: observations are missing because the locations of interest do not have historical data recorded at all (e.g., no sensors have been deployed at those locations). 
For the latter category, recent studies revisit  Kriging~\cite{orignalKriging}. The aim is to generate fine-grained records through coarse-grained observations by inserting derived data for the unobserved locations. The state-of-the-art models~\cite{ignnk,increase} adopt neural networks as a solution. \emph{Both categories assume some data available for the region of interest, shown in Fig.~\ref{fig:problem} (a) and ~\ref{fig:problem} (b).}

Existing studies fail to consider a "continuous" data scarcity scenario, where all locations without any data observations are skewed towards a continuous sub-region (\emph{i.e., the region of interest has no available data}) while the region boundary is adjacent to the locations with data observations. Fig.~\ref{fig:problem} (c) illustrates this scenario, which may occur when (1) sensors are deployed progressively from one region to another (one such scenario has been observed in Hong Kong~\cite{dastnet}), (2) some regions do not have resources to deploy the sensors (e.g., Shanghai traffic congestion data only covers the core city region~\cite{shanghai}), or (3) regions are not willing to open their data (e.g., the Tom-Tom traffic index~\cite{tomtom} does not have data from mainland China). 

To fill this gap, we propose a new problem -- \emph{spatial-temporal forecasting for a region of interest without historical data.}

This new problem is challenging, and existing Kriging-based approaches~\cite{increase,ignnk} do not address the problem directly. This is because the core idea of the Kriging-based approach is to perform data interpolation for the locations without observations based on nearby locations (or time) with observations, which is unavailable under our setting.

For example, an earlier model, IGNNK~\cite{ignnk}, represents the region of interest as a graph and exploits GNNs for Kriging. It reports substantial performance drops in our setting (detailed in Section~\ref{subsec:experiment_results}) because GNNs become ineffective when the local neighbourhood of a location does not have historical observations to help infer the data. The state-of-the-art Kriging model, INCREASE~\cite{increase}, also uses a graph-based region representation. It aggregates information from nearest neighbours in advance and then uses GRUs (which is a type of RNN) to capture the temporal correlation of the data.
This model fails to utilise the global features of the graph as it only considers the nearest neighbours.

We propose a \underline{\emph{s}}patial-\underline{\emph{t}}emporal forecasting model with a \underline{\emph{s}}elective \underline{\emph{m}}asking strategy, named \model, to enable learning from the locations that resemble those in the region of interest and forecasting for the region of interest without observations.
We mask observed data from sub-regions (which are near the region of interest and have historical data) and train \model\ to make predictions for the sub-regions. Then, at testing time, we exploit the similarity between the region of interest and the sub-regions used at training to make predictions for the region of interest. 

Unlike existing studies that use random masking~\cite{ignnk}, \model\ incorporates a \emph{selective masking module} to mask sub-regions that are more similar to the region of interest. This strategy makes it easier for \model\ to extend its forecasting capability from the masked sub-regions to the region of interest.
The selective masking module fuses regional features, road network features and spatial distance to compute similarity scores between the masked sub-region (i.e., masked locations) and the region of interest (i.e., unobserved locations). The similarity scores are normalised into the range of [0,1] and serve as the probabilities for drawing the sub-regions (or locations) to be masked.

Further, we use the observed locations' historical data to generate pseudo-observations for the unobserved locations and the masked locations, which enables computing a temporal similarity based adjacency matrix, i.e., we build links between observed locations and unobserved locations that have high temporal similarities. This helps identify more similar neighbours.

Overall, this paper makes the following contributions:
\begin{itemize}
\item We propose a new spatial-temporal forecasting task - forecasting for regions without historical observations. This task can be applied to addressing issues of unbalanced region development and lack of open data.
\item We design a selective masking module to guide our model \model\ to mask observed locations that have high similarity with the unobserved region to enable \model\ to generalise predictions to the unobserved region.
\item We design an efficient pseudo-observation generating strategy and compute a temporal adjacency matrix based on it to help identify the more informative neighbours and enhance model learning efficacy. 
\item Extensive experiments show that our model outperforms the state-of-the-art model that we adapt to this new problem, in terms of forecasting accuracy. 
\end{itemize}

\section{Related Work}

\subsection{Spatial-temporal Forecasting}
Current state-of-the-art spatial-temporal forecasting models are mainly based on deep neural networks.
DCRNN~\cite{dcrnn} introduces a diffusion convolutional recurrent neural network to model the spatial correlation between locations and adapts gated recurrent units (GRU) to model the temporal dependency.
GRUs and other RNN models have a recurrent structure that suffers in model running time and in the effectiveness of modelling longer sequences. To overcome this limitation, GraphWaveNet~\cite{graphwavenet} utilises 1-D temporal convolutional modules to capture the temporal dependency. In addition, the attention mechanism~\cite{att_1} is widely used in spatial-temporal forecasting~\cite{astgcn, gman, ST-aware, stwave}. 
A series of recent studies further embed heterogeneous relations into adjacency matrices, including temporal similarity~\cite{stsgcn,stfgnn} and embedding similarity~\cite{gamcn}. Some studies~\cite{stssl,pdformer,whdocl,SSTBAN} adopt self-supervised learning to enhance spatial-temporal pattern representations. Meanwhile, DeepSTUQ~\cite{DeepSTUQ} considers prediction uncertainty when forecasting traffic. 
These models assume fully available historical data and suffer in learning capability when data is incomplete.

\subsection{Spatial-temporal Forecasting with Incomplete Data}
\label{subsec:forecasting_data_missing}
Existing spatial-temporal forecasting methods for incomplete data can be divided into two categories from the data perspective: 
\emph{data missing at times} and \emph{data missing at (scattering) locations}.

\emph{Data missing at times}:
Observations at locations of interest are incomplete for all time.
For this category, one dominant class of studies focuses on random or continuous data missing at times caused by harsh environments~\cite{mdgcn,surrounding_imputation_clustering}, e.g., extreme weather or transmission device issues. Generative adversarial networks (GAN) are applied to address this issue~\cite{sa-gain, ge-gan}. Another study~\cite{dastnet} uses transfer learning for settings where sensors have only been deployed for a short period of time  (e.g., 10 days). 

\emph{Data missing at (scattering) locations}:
Some locations of interest do not have observations at all.
Problems considering this setting, i.e., Kriging~\cite{orignalKriging}, aim to impute fine-grained records via coarse-grained records, which is to recover signals for unobserved locations. Gaussian process regression~\cite{gaussian_kriging} is a classic solution, while it suffers from low efficiency and poor scalability. 
Tensor/matrix completion algorithms~\cite{k_matrix_factorization,tensor-factorization,low-rank-kriging} show better efficiency on large datasets.
They combine the low-rank structure and regularisers to maintain local and global consistency. Most tensor/matrix completion algorithms are transductive. They cannot process a new location of interest without re-training. Recent studies~\cite{ignnk,increase} propose inductive structures. For example, IGNNK~\cite{ignnk} utilises the inductive nature of GNNs together with 1-D CNN to recover records for unobserved locations. This model struggles to handle high data missing ratios where there is little information to learn from the neighbourhood. INCREASE~\cite{increase} adopts an RNN for inductive imputation. This method uses heterogeneous relations for more accurate estimation, while it struggles in capturing global spatial-temporal patterns. 

\subsection{Graph Contrastive Learning}
Our proposed \model\ is based on contrastive learning, in particular, graph contrastive learning (GCL), which applies contrastive learning to graph data. The basic idea of contrastive learning is to maximize the similarity between positive samples while minimizing that between negative samples.

A series of studies~\cite{graphcl, gca, JOAO} focuses on graph augmentation modules that generate positive and negative samples. For example, GraphCL~\cite{graphcl} introduces four augmentation methods, such as node dropping and edge perturbation, to create positive graph pairs. Later, GCA~\cite{gca} and JOAO \cite{JOAO} improve the augmentation strategies by taking the node weights and the edge weights into consideration.
Besides, some studies~\cite{infograph,dgi} aim to maximize the mutual information between graph inputs at different scales, e.g., nodes vs. graphs.

A few studies~\cite{whdocl, chang_contrastive} introduce GCL into spatial learning tasks. For example, SARN~\cite{chang_contrastive} learns road network embeddings via contrastive learning,
STGCL~\cite{whdocl} applies GCL to predict traffic flow assuming complete data.
Unlike STGCL, our model can process incomplete data because of the proposed selective masking module that can adaptively augment spatial-temporal graphs based on the heterogeneous similarities between observed and unobserved locations.

\begin{figure*}[h!]
         \centering
  \includegraphics[width=0.9\linewidth]{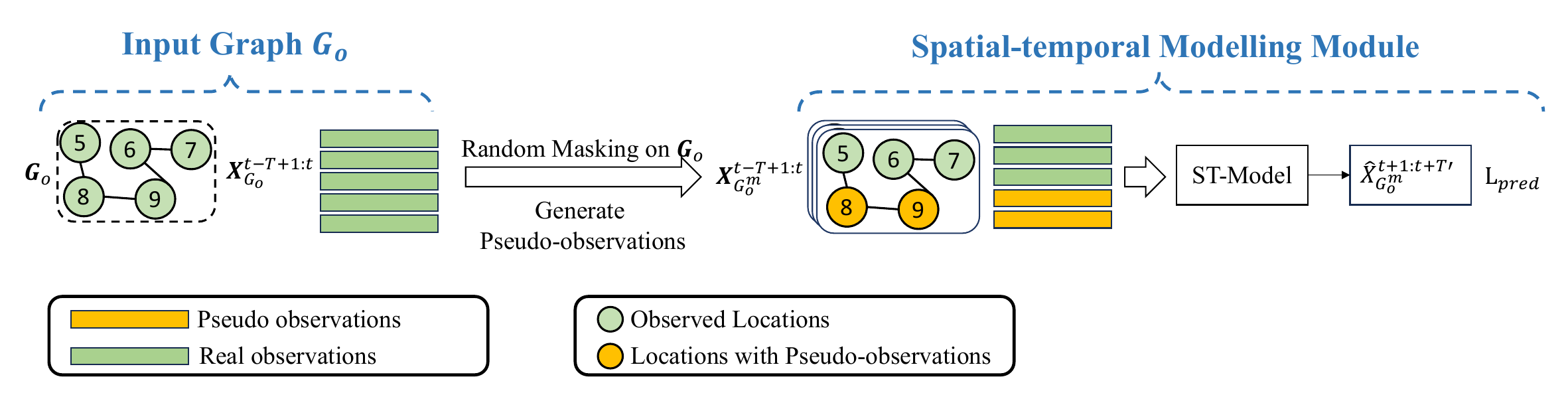}
         \caption{Model architecture of \model-RNC. The model contains a sub-graph masking module and a spatial-temporal modelling module.}
         \label{fig:base_model}
\end{figure*}

\section{Proposed Base Model}
\label{sec:Proposed basic model}
We start with a basic version of our proposed model, named \model-RNC (Fig.~\ref{fig:base_model}). We first define basic concepts and our studied problem. Then, we present our base model and its training and testing procedures. 

\subsection{Concepts and Problem Statement}
 \textbf{Region and Region Graph.} We represent a region as a graph $G=(V, E)$. The set of graph vertices $V$ represents $N$ locations of interest in the region, and the set of graph edges $E$ represents the connection between the locations. 
Graph $G$ has a feature matrix $\textbf{L} \in \mathbb{R}^{N \times F}$ for the locations, where $F$ is the dimensionality of the location features. The features of a location consist of two parts, i.e., region information and road network information, which will be detailed in Section~\ref{subsec:selec_mask}. 

For each location $v_i \in V$, $x_i^t \in \mathbb{R}^C$ represents the \emph{observations} at $v_i$ at time step $t$, where $C$ is the number of different types of observations, e.g., traffic speed, PM2.5, etc.

\textbf{Observed and unobserved regions.} Region graph $G$ can be further divided into two adjacent sub-regions based on the availability of observations for the sub-regions, i.e., \emph{the observed region} and \emph{the unobserved region}. The locations in these two regions are referred to as \emph{observed locations} (i.e., with observations) and \emph{unobserved locations} (i.e., without observations), respectively. We use $R_o$ to denote the region that contains \emph{all and only the} observed locations. Similarly, we use $R_u$ to denote the region that contains \emph{all and only the} unobserved locations. These two regions have no overlap with each other, i.e., $R_o \cap R_u =\phi$.
We use $G_o = (V_o, E_o)$ to represent the graph on the observed locations, where $V_o \subset V$ denotes the set of the observed locations, and $E_o \subset E \cap (V_o\times V_o)$ denotes the set of the edges over $V_o$. We use $N_o$ to denote the size of $V_o$.
Similarly, we define the graph on the unobserved locations, which is denoted as $G_u$, and the nodes $V_u$ (and its size $N_u$) and edges $E_u$ on this graph.
Note that, $N = N_o+N_u$; 
$\textbf{X}_{G_o}^{t}=(x_{1}^{t},...,x_{N_o}^{t}) \in \mathbb{R}^{N_o \times C}$ denotes the observations of the observed locations in $G_o$ at time steps $t$; and $\hat{\textbf{X}}_{G_u}^{t}=(\hat{x}_{1}^{t},...,\hat{x}_{N_u}^{t}) \in \mathbb{R}^{N_u \times C}$ denotes estimated values for the unobserved locations in $G_u$ at time step $t$. 

\textbf{Problem definition.}
Given a region graph $G$ as defined above, with location features $\textbf{L}$ and past observations on the observed locations for a time window of length $T$, we aim to learn a function $f$ to predict the values for the unobserved locations over the next $T'$ time steps:

\begin{equation}\label{eq:task_definition}
    \hat{\textbf{X}}_{G_u}^{t+1},...,\hat{\textbf{X}}_{G_u}^{t+T'} = f(\textbf{X}_{G_o}^{t-T+1},...,\textbf{X}_{G_o}^{t}; G; \textbf{L})
\end{equation}

\subsection{Overview of Our Base Model}
We next present a basic model (i.e., \model-RNC in Section~\ref{subsubsec:ablation_study}) that directly combines \emph{spatial-temporal modelling} with \emph{random sub-graph masking} to forecast for regions without observations as Fig.~\ref{fig:base_model} shows. The main idea of \model-RNC is to learn a model that can forecast observations (e.g., traffic speed or PM2.5) for masking sub-graphs and to extend this capacity to forecast for unobserved locations. 

We denote the full graph with those unobserved locations as $G$. We mask a subset of the locations of $G_o$ to generate a masked view $G_o^m$ (Section~\ref{subsec:sub_graph_mask}). 
Following previous studies~\cite{dcrnn,gcnode}, we use temporal similarity-based adjacency matrix and spatial-based adjacency matrix for spatial correlation modelling. To compute the temporal similarity for the unobserved locations, we first compute pseudo-observations for all unobserved locations. 
Then, we use dynamic time warping (DTW)~\cite{dtw,stfgnn} to compute temporal similarities among all observed locations, and the temporal similarities between the observed and the unobserved locations. Besides, for the masked locations in each model training epoch, we compute pseudo-observations for them and compute the temporal similarities between masked locations and observed locations. After these steps, we obtain $\textbf{X}_{G_o^m}^{t-T+1:t}$ and $\textbf{X}_{G^m}^{t-T+1:t}$ for training and testing, respectively, where both $\textbf{X}_{G_o^m}^{t-T+1:t}$ and $\textbf{X}_{G^m}^{t-T+1:t}$ contain the pseudo-observations for the masked or unobserved locations in $G_o$ or $G$. 

We feed $\textbf{X}_{G_o^m}^{t-T+1:t}$ into a \emph{spatial-temporal modelling module} to generate the prediction result $\textbf{X}_{G_o^m}^{t+1:t+T'}$ (Section~\ref{subsec:st_model}) and compute the mean squared error between the prediction and the ground truth as prediction loss to optimise the spatial-temporal model. After the model is trained, we feed $\textbf{X}_{G^m}^{t-T+1:t}$ into the model to obtain predictions for the unobserved locations (Section~\ref{subsec:model_train_test}). 

\subsection{Sub-graph Masking}
\label{subsec:sub_graph_mask}
\model-RNC learns to predict values for masked locations during training and then extends this capability to predict values for unobserved locations at testing. To simulate the setting of a continuous region without data observations, which we focus on, we mask the sub-graph formed by each selected location and its 1-hop neighbours instead of a set of scattering locations.

\textbf{Defining a sub-graph.} The sub-graph of an observed location is formed by its 1-hop neighbours. We compute a location's 1-hop neighbours based on a spatial adjacency matrix $\textbf{A}_{sg}$, which is defined by Eq.~\ref{eq:A_sg}, where $\epsilon_{sg}$ is a hyper-parameter, and the $dist(c_i,c_j)$ denotes the distance between locations $i$ and $j$ ($c_i$ and $c_j$ are their geo-coordinates.) We use Euclidean distance in this paper for efficiency considerations, though road network distance can also be used.

\begin{equation}\label{eq:A_sg}
  A_{sg,ij}=\begin{cases}
  1 & exp(-\frac{dist(c_i,c_j)^2}{\sigma^2}) \geq \epsilon_{sg}, \\
  0 & \text{otherwise}.
\end{cases}
\end{equation}

\textbf{Sub-graph masking.} We use a masking ratio $\delta_m$ to define the percentage of observed locations to be masked. The number of locations masked is expected to be $N_o \cdot \delta_m$. Since the sub-graph of each location may have a different size, \model-RNC iteratively and randomly selects a location and masks the location and its 1-hop neighbours until the number of masked locations reaches $N_o \cdot \delta_m$. 

\begin{figure}[h!]
         \centering
  \includegraphics[width=\linewidth]{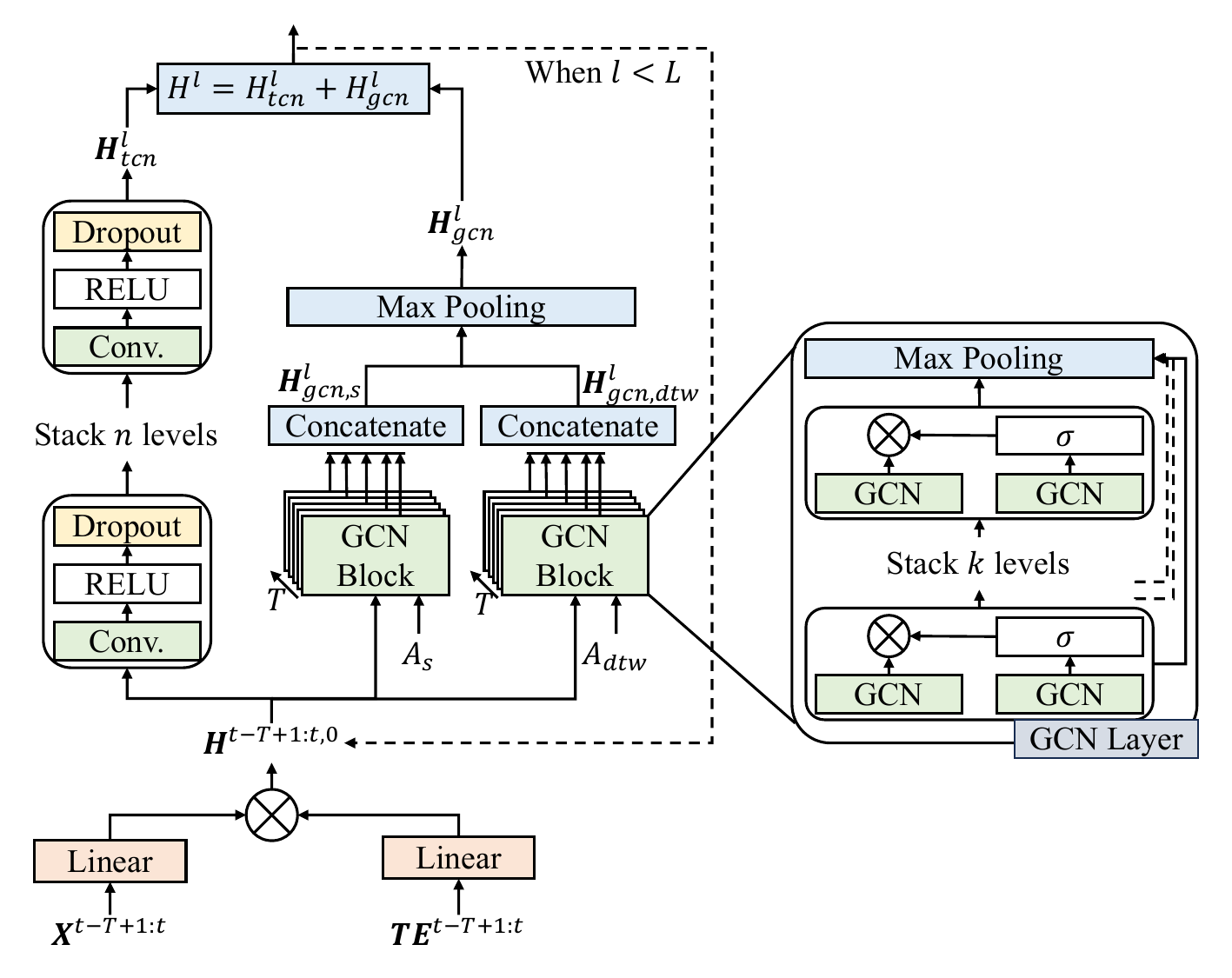}
         \caption{The structure of the spatial-temporal model}
         \label{fig:st_model}
\end{figure}

\subsection{Spatial-temporal Modelling}
\label{subsec:st_model}
The spatial-temporal modelling module of our base model \model-RNC contains the 1-D convolution networks for temporal correlation modelling and the graph convolutional networks (GCNs) for spatial correlation modelling. The spatial-temporal modelling module stacks multiple blocks to compute the final output. Fig.~\ref{fig:st_model} shows the structure of the spatial-temporal modelling module and details the $l$th block of the module. Each block contains a temporal correlation modelling module and a spatial correlation modelling module. These two modules are parallel in each block.
We first describe the input features and the adjacency matrices used in the spatial-temporal modelling module. Then, we detail the temporal and spatial correlation modelling modules.

\subsubsection{Input Features and Adjacency Matrices}
The input features of \model-RNC consist of historical observations for the observed locations, pseudo-observations for the unobserved and the masked locations, and temporal attention which is used to indicate the time of a day. 
Besides, we compute two types of adjacency matrices for the GCNs in our model.

We compute pseudo-observations for the $i$-th unobserved or masked location from real observations $x_{i}^{t}=\sum_{j \in N_o}\alpha_{i,j}x_{j}^{t}$. 
The weight for each observed location is determined by its spatial distance to location $i$, as defined by  
Eq.~\ref{eq:weight_4_pseudo_obv}:
\begin{equation}\label{eq:weight_4_pseudo_obv}
  \alpha_{i,j} = \frac{{dist(c_i,c_j)^{-1}}}{\sum_{l \in N_o}dist(i,l)^{-1}}
\end{equation}
 
This step can introduce more information into an unobserved or masked location based on that of its neighbours. 

Then, we build a temporal similarity-based adjacency matrix.  
We follow a prior work~\cite{stfgnn} and adopt DTW to compute the temporal similarity. Since pseudo-observations can be regarded as real observations with noises, we only build links among the observed locations and from observed locations to  unobserved (masked) locations (i.e., unobserved locations cannot send messages to observed locations directly during GCN training). This way, we avoid polluting the embeddings of the observed locations by those of the unobserved (masked) locations. 
We compute $q_{kk}$ and $q_{ku}$ most similar pairs of observed locations and pairs of observed and unobserved (or masked) locations, respectively. We establish a temporal similarity-based adjacency matrix ($\textbf{A}_{dtw}^{train} \in \mathbb{R}^{N_o \times N_o}$ in the training process and $\textbf{A}_{dtw} \in \mathbb{R}^{N \times N}$ in the testing process) by assigning an edge weight of 1 for these pairs, and 0 for the rest of the location pairs. 
Since the locations are masked dynamically in each training epoch, $\textbf{A}_{dtw}^{train}$ is updated in each model training epoch.

Further, we use temporal attention to capture periodic patterns, which can significantly impact the observed values of spatial-temporal data, e.g., rush hours. Given the length of a time interval that the spatial-temporal observations are recorded (e.g., 5 minutes), 
we can compute the number of intervals in a day, denoted as $Td$. Now each observation interval in a day gets an interval ID in $[0, Td-1]$. Given an input of length $T$, we compute a time-of-day embedding $TE \in \mathbb{R}^T$, which stores the interval IDs in the input time window.  For example, $TE = [0, 1, 2, 3]$ represents an input observation sequence that starts at the first interval of a day and ends at the fourth interval of a day.

To attach the time embedding $TE$ to the model input, we first project $\textbf{TE}^{t-T+1:t} \in \mathbb{R}^{N_o \times T \times 1}$ and the input features $\textbf{X}^{t-T+1:t} \in \mathbb{R}^{N_o \times T \times C}$ into the same latent space and then multiply them as formulated by Eq.~\ref{eq:tattt}:

\begin{equation}\label{eq:tattt}
  \textbf{H}^{t-T+1:t,0} = \phi_1(\textbf{X}^{t-T+1:t})\otimes \phi_2(\textbf{TE}^{t-T+1:t})
\end{equation}
Here, $\textbf{X}^{t-T+1:t} \in \mathbb{R}^{N_o \times T\times C}$ is the input observation sequence of the observed graph (i.e., $G_o$ or $G_o^m$), while  $\textbf{TE}^{t-T+1:t}$ is the corresponding time embedding; $\phi_1(\cdot)$ and $\phi_2(\cdot)$ are linear functions that project the input observation sequence and time embedding into the same latent space for element-wise multiplication. We now obtain the features $\textbf{H}^{t-T+1:t,0} \in \mathbb{R}^{N_o \times T\times C'}$, which serve as the inputs to the spatial-temporal model.

\subsubsection{Temporal Correlation Modelling}
1-D convolution neural networks have shown strong performance in temporal feature modelling. We adopt 1-D dilated convolution neural networks to embed the temporal features. For ease of presentation, we simplify the notation of the input features of the first layer of the spatial-temporal model, $\textbf{H}^{t-T+1:t,0}$, to $\textbf{H}^{0}$.

\begin{equation}\label{eq:tch}
  \textbf{H}^{l}_{tcn} = *_{d^{l}}\sigma(W^{l}_{d^l_j}\textbf{H}^{l-1}) 
\end{equation}
Here, $\textbf{H}^{l-1}\in \mathbb{R}^{N \times T \times C'}$ is the output of the previous layer;  $\textbf{H}^{l}_{tcn} \in \mathbb{R}^{N \times T \times C'}, l=1, 2, \ldots, L$ is the output of the $l$-th layer's 1-D temporal convolution networks;  $*_{d^{l}}$ means to stack 1-D dilated temporal convolution networks, where $d^{l}_j$ represents the exponential dilation rate, $d^{l}_j=2^{j}$. To keep the same dimensionality for the time sequence representation, we use zero-padding. Function  $\sigma(\cdot)$ is the activation function (e.g., \emph{ReLU} or \emph{sigmoid}).

\subsubsection{Spatial Correlation Modelling} We use a graph convolutional network (GCN) to model spatial correlations. The basic idea of GCN is to aggregate features from neighbours:
\begin{equation}\label{eq:gcn}
  GCN(\emph{\textbf{A, H}}) = \widetilde{\textbf{D}}^{-1/2}\widetilde{\textbf{A}}\widetilde{\textbf{D}}^{-1/2}\textbf{ZW}
\end{equation}
where $\widetilde{{\textbf{A}}}={\textbf{A}}+{\textbf{I}}$, and $\widetilde{\textbf{D}}$ is a diagonal matrix. Matrix $\textbf{Z}\in \mathbb{R}^{N \times C}$ is the input graph node features. 
Matrix $\textbf{W} \in \mathbb{R}^{C \times C'}$ contains the parameters to be learned by the model, where $C$ is the input dimensionality and $C'$ is the output dimensionality. 
Next, we define GCN layers with two parallel GCNs, denoted as $GCNL$:
\begin{equation}\label{eq:gcnl}
  GCNL(\textbf{A, Z}) = GCN(\textbf{A},\textbf{Z}) * \emph{sigmoid}(GCN(\textbf{A},\textbf{Z}))
\end{equation}

\begin{figure*}[t]
         \centering
  \includegraphics[width=\linewidth]{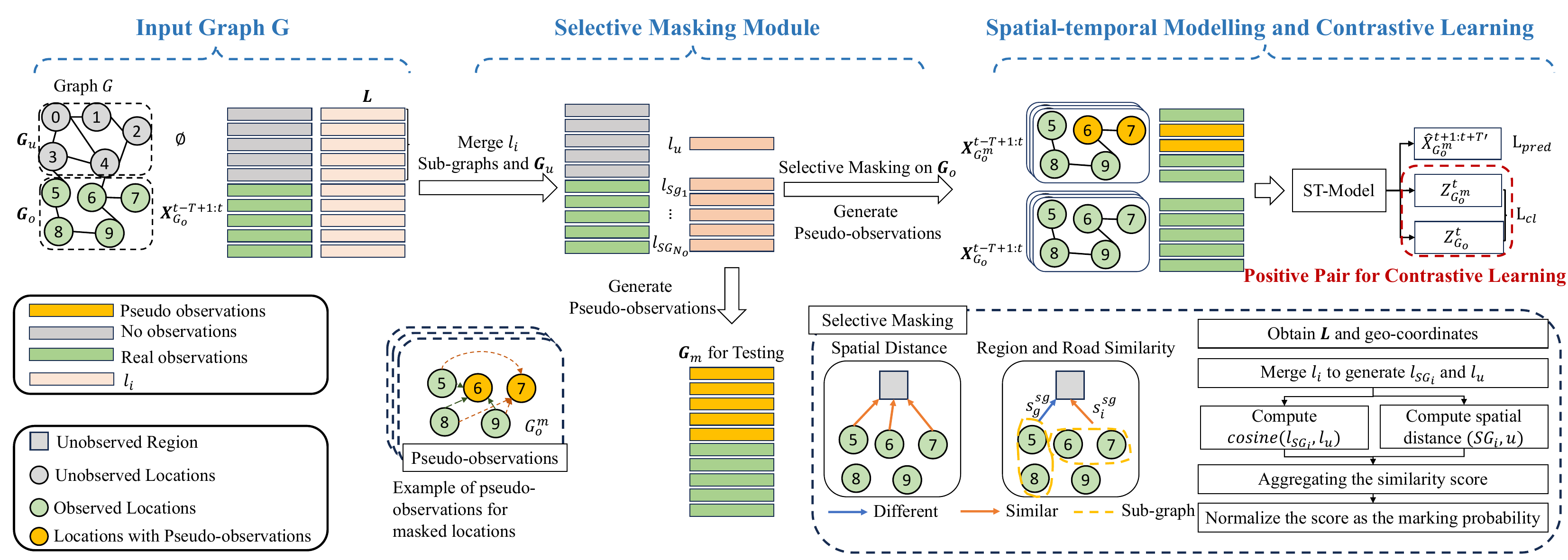}
         \caption{Model architecture of \model. The model contains three main parts. (1)~\emph{The selective masking module} leverages the regional and road network representations and the spatial distances to compute the similarity between observed locations (i.e., their sub-graphs) and the unobserved region. Masking probabilities are assigned based on the similarity scores. (2)~\emph{The contrastive learning module} guides \model\ to make similar predictions for location graphs with complete data and graphs with incomplete data.
         (3)~\emph{The spatial-temporal modelling module} (as described in Section~\ref{subsec:st_model}) utilises GCNs and 1-D TCNs to model spatial and temporal features, together with a contrastive learning loss to optimise the model. To enhance model performance, \model\ generates pseudo-observations for unobserved locations and computes a temporal similarity-based adjacency matrix. During the testing process, \model\ fills unobserved locations with pseudo-observations and then feeds the graph into ST-Model to obtain the prediction results.}
         \label{fig:framework}
\end{figure*}

We stack GCN layers to build a GCN block. The output of each GCN layer is the input of the next GCN layer, as shown in Eq.~\ref{eq:stachgcn}, where $q \in [1, k]$. 
The first layer input is $\textbf{H}^{l,t-p,0}_{gcn,r} = \textbf{H}^{l-1,t-p}$, where $p \in [T-1, 0]$ and $r \in \{s,dtw\}$ represents two types of adjacency matrices (i.e., spatial proximity-based matrix and a temporal similarity-based matrix).
\begin{equation}\label{eq:stachgcn}
  \textbf{H}^{l,t-p,q}_{gcn,r} = GCNL_{l,q-1}(\textbf{A}_r,\textbf{H}^{l,t-p,q-1}_{gcn,r})
\end{equation}
We use $max(\cdot)$ to aggregate the outputs of the GCN layers to obtain the output of $l$-th GCN block (Eq.~\ref{eq:gcnb})
\begin{equation}\label{eq:gcnb}
  \textbf{H}^{l,t-p}_{gcn,r} = max_{q=1,...,k}(GCNL_{l,q}(\textbf{A}_r,\textbf{H}^{l,t-p,q}_{gcn,r}))
\end{equation}
Then, we concatenate the output from each time slot:
\begin{equation}\label{eq:gcncat}
  \textbf{H}^{l}_{gcn,r} = ||_{p\in [T-1,0]}\textbf{H}^{l,t-p}_{gcn,r}
\end{equation}
After that, we use $max(\cdot)$ again to aggregate the outputs corresponding to the two different adjacency matrices, as shown in Eq.~\ref{eq:gcnoutput}, to obtain the output of the $l$-th layer:
\begin{equation}\label{eq:gcnoutput}
  \textbf{H}^{l}_{gcn} = max_{r \in \{s,dtw\}}(\textbf{H}^{l}_{gcn,r})
\end{equation}

We follow previous studies~\cite{dcrnn,gcnode} and adopt Eq.~\ref{eq:A_sg} with different threshold - $\epsilon_s$ to compute the spatial-based adjacency matrix. 
Meanwhile, we follow another work~\cite{stfgnn} and adopt DTW~\cite{dtw} to compute $\textbf{A}_{dtw}$, as described earlier. 

We combine the output of TCN and GCN to obtain the output of the $l$-th layer:
\begin{equation}\label{eq:outputl}
  \textbf{H}^{l} = \textbf{H}^{l}_{gcn} + \textbf{H}^{l}_{tcn}
\end{equation}

Finally, we obtain the output $\textbf{H}^{t+1:t+T',L}$ at the $L$-th layer following the steps described above and linear functions to project $\textbf{H}^{t+1:t+T',L}$ to a lower dimension (as Eq.~\ref{eq:output} shows).
\begin{equation}\label{eq:output}
  \hat{\textbf{X}}^{t+1:t+T'} = \sigma(\phi_4(\sigma(\phi_3(\textbf{H}^{t+1:t+T',L}))))
\end{equation}
Here, $\phi_3$ and $\phi_4$ are linear functions, and $\sigma$ is an activation function. $\hat{\textbf{X}}^{t+1:t+T'} \in \mathbb{R}^{N_o\times T' \times C}$ represents the prediction values.

\subsection{Model Training and Testing}
\label{subsec:model_train_test}
\textbf{Model Training:}
We obtain predicted values $\hat{\textbf{X}}^{t+1:t+T'}_{G_o^m}$ on the graph view $G_o^m$ that is generated by sub-graph masking. Then, we compute the mean squared error between the prediction values $\hat{x}_{i}^{t+p'}$ and the ground truth $x_{i}^{t+p'}$ as the prediction loss (Eq.~\ref{eq:pred_loss}). 
\begin{equation}\label{eq:pred_loss}
  L_{pred} = \frac{1}{N_{o}T}\sum_{i=1}^{N_o}\sum_{p'=1}^{T}||\hat{x}_{i}^{t+p'} - x_{i}^{t+p'}||_2^2
\end{equation}

\textbf{Model Testing:} During model testing, we first compute pseudo-observations for the unobserved locations, and let the graph $G$ with pseudo-observations be $G_m$. Then, we build the temporal similarity-based adjacency matrix utilising these pseudo-observations. After that, we feed the features $\textbf{X}_{G_m}^{t-T+1:t}\in \mathbb{R}^{N\times T\times C}$ into the trained model to produce the predicted observations $\hat{\textbf{X}}_{G_m}^{t+1:t+T'}$ for the unobserved locations.

\section{Proposed Full Model}
Section~\ref{sec:Proposed basic model} introduced our base model. In this section,  we will introduce two modules - \emph{selective masking module} and \emph{contrastive learning module} that enhance our proposed model performance. These two modules together with our base model \model-RNC form our full model \model. Figure~\ref{fig:framework} shows its overall structure. 

Recall that our core idea is to learn a model that can extend the forecasting capability for masked locations to unobserved locations. 

The generalisability of \model\ on performance on the full graph $G$ depends on the similarity between the masked locations in $G_o^m$ and the unobserved locations in $G$. To mask the locations that have higher similarity with the unobserved locations, we propose a \emph{selective masking module}, to enhance the \emph{sub-graph masking}, exploiting the similarity between the observed locations and the unobserved locations to help forecast the values (e.g., traffic speed or PM2.5) at the unobserved locations. This module leverages the regional information and the road network information surrounding the locations, as well as the spatial distance to compute the masking probabilities of the locations. We mask the locations based on such probabilities at each model training epoch to generate $G_o^m$. This module can guide \model\ to learn to predict values for the locations that have higher similarities with the unobserved locations, thus enhancing the generalisability of the model.

We further design a contrastive learning module that takes a graph contrastive learning-based approach and constructs two views of the graph - one view contains complete spatial-temporal data (the original view), and the other view contains incomplete spatial-temporal data (the augmentation view). The view with complete data is used to guide the prediction for the view with incomplete data. The augmentation view $G_o^m$ is generated by the selective masking module. Using contrastive learning, we learn a model that generates similar predictions for both graph views. The trained model is then applied on the full graph $G$ to make predictions for the unobserved locations. 

We feed $\textbf{X}_{G_o}^{t-T+1:t}$ and $\textbf{X}_{G_o^m}^{t-T+1:t}$ into the proposed \emph{spatial-temporal modelling module} (described in Section~\ref{subsec:st_model}) to obtain the graph representations $\textbf{Z}_{G_o}^{t+T'}$ and $\textbf{Z}_{G_o^m}^{t+T'}$ for contrastive learning and generating the prediction result $\textbf{X}_{G_o^m}^{t+1:t+T'}$. 

When the model is trained, we feed $\textbf{X}_{G^m}^{t-T+1:t}$ into the model to obtain predictions for the unobserved locations (Section~\ref{subsec:model_train_test}).

\subsection{Selective Masking}
\label{subsec:selec_mask}
\model\ learns to predict values for masked locations during training, and then extends this capability to predict values for unobserved locations at testing.
Intuitively, the higher the similarity between the masked locations and unobserved locations, the easier it is for the trained model to make predictions for the unobserved locations. 
We compute similarities between those 1-hop sub-graphs (defined in Section~\ref{subsec:sub_graph_mask}) and the unobserved region. After that, we use our proposed selective masking module to guide \model\ to mask a sub-region formed by the sub-graphs of observed locations that are the most similar to the unobserved region. Heuristically, this strategy leads to more accurate forecasting results for the locations in the unobserved region.

\begin{table*}[!h]
\centering
  \caption{Categories of Points of Interest}
  \label{tab:poi_categories}
\begin{tabular}{p{1.3cm}|p{6.4cm}|p{1.3cm}|p{6.4cm}}
  \hlineB{3}
  Categories & Subcategories & Categories & Subcategories\\
  \hline \hline
  \#1 & university, college, school, kindergarten, research institute, language school, childcare &
  \#2 & commercial, office, studio\\
  \hline
  \#3 & retail, supermarket &
  \#4 & hotel, motel, guest house, hostel\\
  \hline
  \#5 & arts centre, library, events venue, community centre, conference centre, theatre, exhibition centre, planetarium, music venue, gallery, artwork, museum, social centre, zoo, aquarium, theme park &
  \#6 & clinic, hospital, veterinary, pharmacy, doctors, nursing home, dentist, social facility, baby hatch\\
  \hline
  \#7 & bridges &
  \#8 & cinema \\
  \hline
  \#9 & fountain, garden, park, trampoline park, water park, ranger station, dog park, viewpoint, nature reserve, attraction, bbq &
  \#10 & casino, gambling, nightclub, stripclub, dance\\
  \hline
  \#11 & church, chapel, cathedral, kingdom hall, monastery, mosque, presbytery, religious, shrine, synagogue, temple, place of worship &
  \#12 & cafe, ice cream, restaurant, pub, bar, food court, fast food, biergarten\\
  \hline
  \#13 & parking, parking entrance, parking exit, parking space, carport, motorcycle parking &
  \#14 & taxi, bus station, transportation, stop position, stop area, train station, platform, station\\
  \hline
  \#15 & warehouse &
  \#16 & industrial\\
  \hline
  \#17 & residential, apartment, apartments &
  \#18 & construction\\
  \hline
  \#19 & marketplace &
  \#20 & caravan site, camp site, camp pitch, picnic site, picnic table\\
  \hline
  \#21 & pitch, sports centre, sports hall, stadium, swimming area, swimming pool, track, grandstand, pavilion, riding hall, sports, fitness centre, fitness station &
  \#22 & civic, government, public\\
  \hline
  \#23 & fuel, car wash, car repair, vehicle inspection, car rental, car sharing &
  \#24 & atm, bank, bureau de change\\
  \hline
  \#25 & boat rental, ferry terminal, boat sharing  &
  \#26 &barn, conservatory, cowshed, farm auxiliary, greenhouse, slurry tank, stable, sty\\
\hlineB{3}
\end{tabular}
\end{table*}

\textbf{Sub-graph representation.} To measure the similarity between a sub-graph and the region formed by the unobserved locations, we need to first compute a representation for the sub-graph of each observed location. We form such a representation (i.e., an embedding) with three components:
\begin{enumerate}
\item \emph{Point of interest (POI) features}. For each observed location, we draw a circle centered at the location with radius $r$ (a system parameter) and collect all POIs inside the circle from OpenStreetMap~\cite{osm}. We classify the POIs into $\Gamma$ categories (cf. Table~\ref{tab:poi_categories}). The POI feature component of the sub-graph embedding, denoted by $l_{i}^{poi} \in \mathbb{R}^{\Gamma}$, is a vector that keeps a count of the POIs of each category. We further obtain the number of floors of the buildings and the areas of the parks in the circular region from OpenStreetMap~\cite{osm} to indicate the prosperity of the sub-graph, as $l_{i}^{scale} \in \mathbb{R}^{1}$. 
For example, a sub-graph (i.e., a local region) with a 60-level building is more prosperous than a sub-graph with only a 4-level building.
We concatenate $l_{i}^{poi}$ and $l_{i}^{scale}$ to obtain a sub-graph's region embedding, denoted as $l_{i}^{region} = [l_{i}^{poi}||l_{i}^{scale}] \in \mathbb{R}^{{\Gamma}+1}$. Here, $||$ denotes concatenation.

\item \emph{Road network features.} We select the nearest road of the location. To represent the road network corresponding to the sub-graph, we use a 4-dimensional vector $l_{i}^{road} \in \mathbb{R}^{4}$ where the dimensions represent \emph{highway\_level, maxspeed, is\_oneway} and \emph{number of lanes}.

\end{enumerate}
 
Finally, we concatenate the regional representation and the road network representation of location $i$ to form its embedding, i.e., $l_i = [l_{i}^{region} || l_{i}^{road}] \in \mathbb{R}^{\Gamma+5}$. The embedding of the sub-graph of location $i$, denoted by $l_{SG_i}$, is computed as the average embedding of the embeddings of all locations in the sub-graph, i.e., $l_{SG_i}=1/|V_{SG_i}|\sum_{j\in V_{SG_i}}{l}_{j}$.

\textbf{Similarity between sub-graphs and the unobserved region.} Following the same strategy, we can compute an embedding $l_u$ for the full unobserved region by averaging the embeddings of all unobserved location. Then, the similarity between the sub-graph of location $i$ and the unobserved region is computed as the cosine similarity of the two embeddings, i.e., $s^{sg}_i=cosine(l_{SG_i},l_u)$), combined with the spatial proximity $sp_i^{sg}=1/dist(c_i,c_u)$ to guide the masking process. 

We compute the embedding similarities between all sub-graphs and the unobserved region, denoted as $S^{sg} = [s_1^{sg},...,s_{N_o}^{sg}]$, and the spatial proximity $SP^{sg}=[sp_1^{sg},...,sp_{N_o}^{sg}]$.

\textbf{Sub-graph masking.} We use a masking ratio $\delta_m$ to define the percentage of observed locations to be masked. Since the sub-graph of each location may have a different size, we compute the average size of all sub-graphs, denoted as $\delta_s = \frac{1}{N_o}\sum_{i\in N_o}|V_{SG_i}|$. If we mask sub-graphs with the same probability $\delta_{ms} = \delta_m/\delta_s$, the final number of locations masked is expected to be $N_o \cdot \delta_m$. Since we want to use similarity to guide \model\ to mask the observed locations, we combine the similarities $S^{sg}$, the spatial proximity $SP^{sg}$ and the sub-graph masking ratio $\delta_{ms}$ to compute the masking probability for each location as Eq.~\ref{eq:norm_sim} shows. This equation normalises $S^{sg}$ and $SP^{sg}$ to make them contribute equally.
\begin{equation}
\label{eq:norm_sim}
  P = (S^{sg} \cdot \frac{\delta_{ms}}{\frac{1}{N_o}\sum_{i \in N_o}s_i^{sg}} + SP^{sg} \cdot \frac{\delta_{ms}}{\frac{1}{N_o}\sum_{i \in N_o}sp_i^{sg}})/2
\end{equation}
The sub-graph size and the size of graph $G$ impact the value $p_i \in P$. When they are large, $p_i$ can become very small, such that all sub-graphs have very close probability values. 
To address this issue, we only keep the top-$K$ most similar sub-graphs and set the similarity values as 0 for the remaining sub-graphs, where $K$ is a hyper-parameter. This strategy reduces the number of sub-graphs that can be masked in a graph. Then, we mask the sub-graphs more similar to the unobserved region, based on a masking probability $\rho_i$, drawn from a Bernoulli distribution $\rho_i\sim Bern(p_i)$ to generate a graph with masked locations ($G_o^m$).

\subsection{Graph Contrastive Learning}
\label{subsec:graph_cl}
Based on Section~\ref{subsec:selec_mask}, we use $G_o$ (i.e., the graph with complete data) to generate a graph $G_o^{m}$ with incomplete data (i.e., a graph with masked locations). Graph $G_o^{m}$ can be regarded as $G_o$ with perturbations (i.e., $G_o^{m}$ and $G_o$ are two views of the observed graph, and $G_o^{m}$ is an augmentation of $G_o$). To guide \model\ to produce similar predictions on $G_o^m$ and $G_o$, we apply contrastive learning in the training process on these two views of the graph.

\model\ adopts graph-level contrastive learning.
We use the original graph $G_o$ to explain our graph representation generating steps.
First, $[\textbf{X}_{G_o}^{t-T+1}, \ldots, \textbf{X}_{G_o}^{t}]$ is inputted into the spatial-temporal model (as described in Section~\ref{subsec:st_model}) to generate an output for each time slot, denoted as $\textbf{H}_{G_o}^{t:t+T',L}$, where $L$ is the number of layers in the spatial-temporal model. 
Then, \model\ takes the last layer output of the spatial-temporal model for the last time step, i.e., $\textbf{H}_{G_o}^{t+T',L}$, to obtain a graph representation $\textbf{Z}_{G_o}^{t+T'}$. We aggregate all locations' representations and project them into a new latent space, formulated as Eq.~\ref{eq:graph_rep}, where $\phi(\cdot)$ is a \emph{linear function}. 
We follow the same steps to generate the representation $\textbf{Z}_{G_o^{m}}^{t+T'}$ of $G_o^{m}$.

\begin{equation}\label{eq:graph_rep}
  \textbf{Z}_{G_o}^{t+T'} = \phi(ReLU(\phi(\sum_{i \in N_o}h_{i,G_o}^{t+T',L})))
\end{equation}

A batch of $M$ input time windows are sampled at training, which form $2M$ representations, where $(\textbf{Z}_{G_o}^{t+T'},\textbf{Z}_{G_o^{m}}^{t+T'})$ is a positive pair (i.e., graph $G_o$ and graph $G_o^m$ from the same time slot form a positive pair). Negative Paris are generated from  the other $M-1$ graphs in the batch (i.e., graph $G_o$ and graph $G_o^m$ from different time slots in a batch form negative pairs), denoted as $(\textbf{Z}_{G_o}^{t+T'},\textbf{Z}_{G_o^{m}}^{t'+T'})$. 
We adopt the contrastive loss to maximise the mutual information of the sample pairs:
\begin{equation}\label{eq:cl_loss}
  L_{cl} = -\frac{1}{M}\sum_{t \in M}log\frac{exp(sim(\textbf{Z}_{G_o}^{t+T'},\textbf{Z}_{G_o^{m}}^{t+T'})/\tau)}{\sum_{t'\in M, t' \neq t}exp(sim(\textbf{Z}_{G_o}^{t+T'},\textbf{Z}_{G_o^{m}}^{t'+T'})/\tau)}
\end{equation}

The final loss function to optimise \model\ is:
\begin{equation}\label{eq:pred_loss}
  L = L_{pred} + \lambda L_{cl}
\end{equation}
where $\lambda$ is a coefficient to balance the prediction loss and the contrastive learning loss.

\section{Experiments}
\subsection{Experimental Setup}
\subsubsection{Datasets.} 
\begin{figure*}[!h]
     \centering
     \begin{subfigure}[b]{0.19\textwidth}
         \centering
        \includegraphics[width=\textwidth]{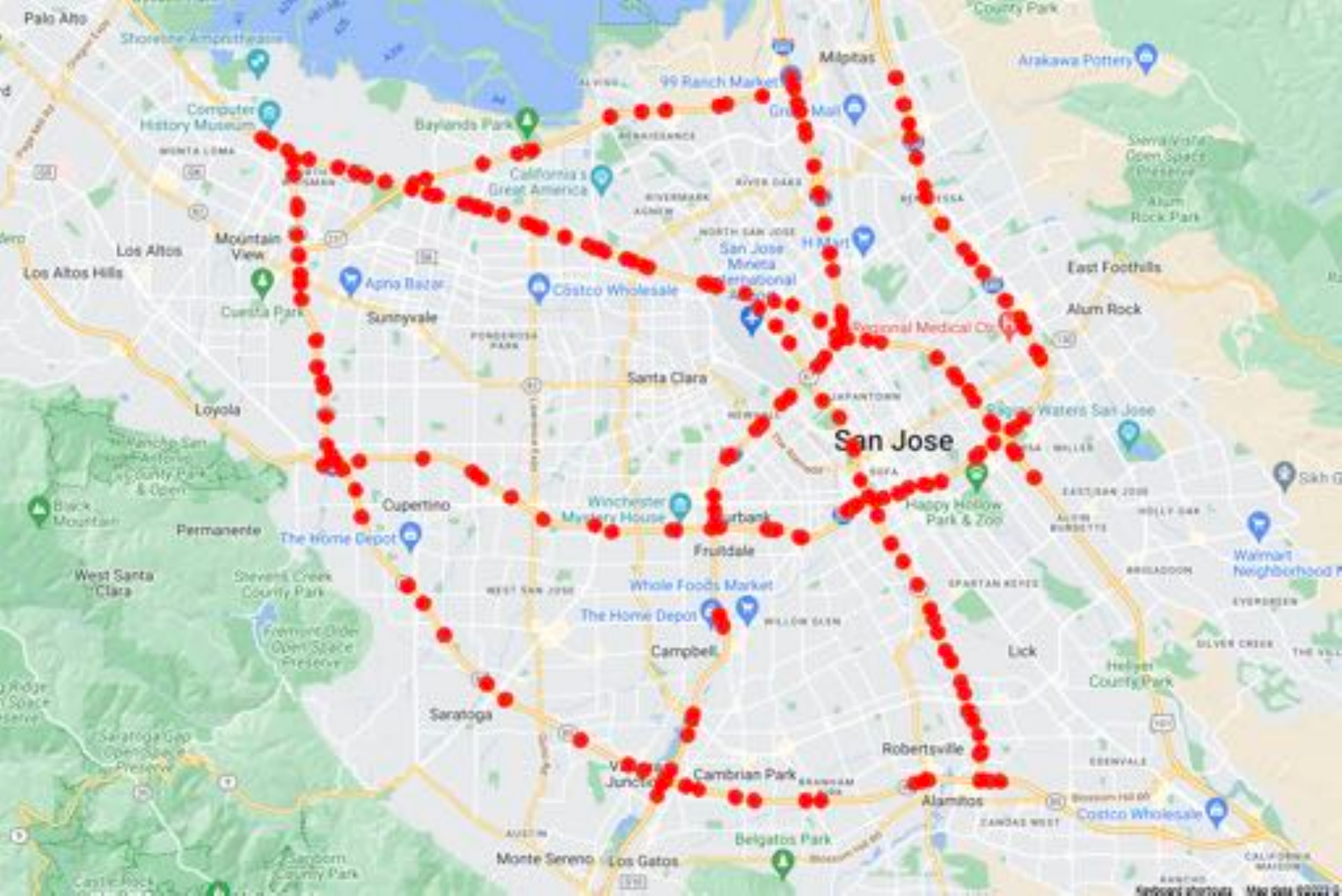}
         \caption{PEMS-Bay}
         \label{fig:sensor_all_bay}
     \end{subfigure}
     \hspace{-1mm}
     \begin{subfigure}[b]{0.19\textwidth}
         \centering
         \includegraphics[width=\textwidth]{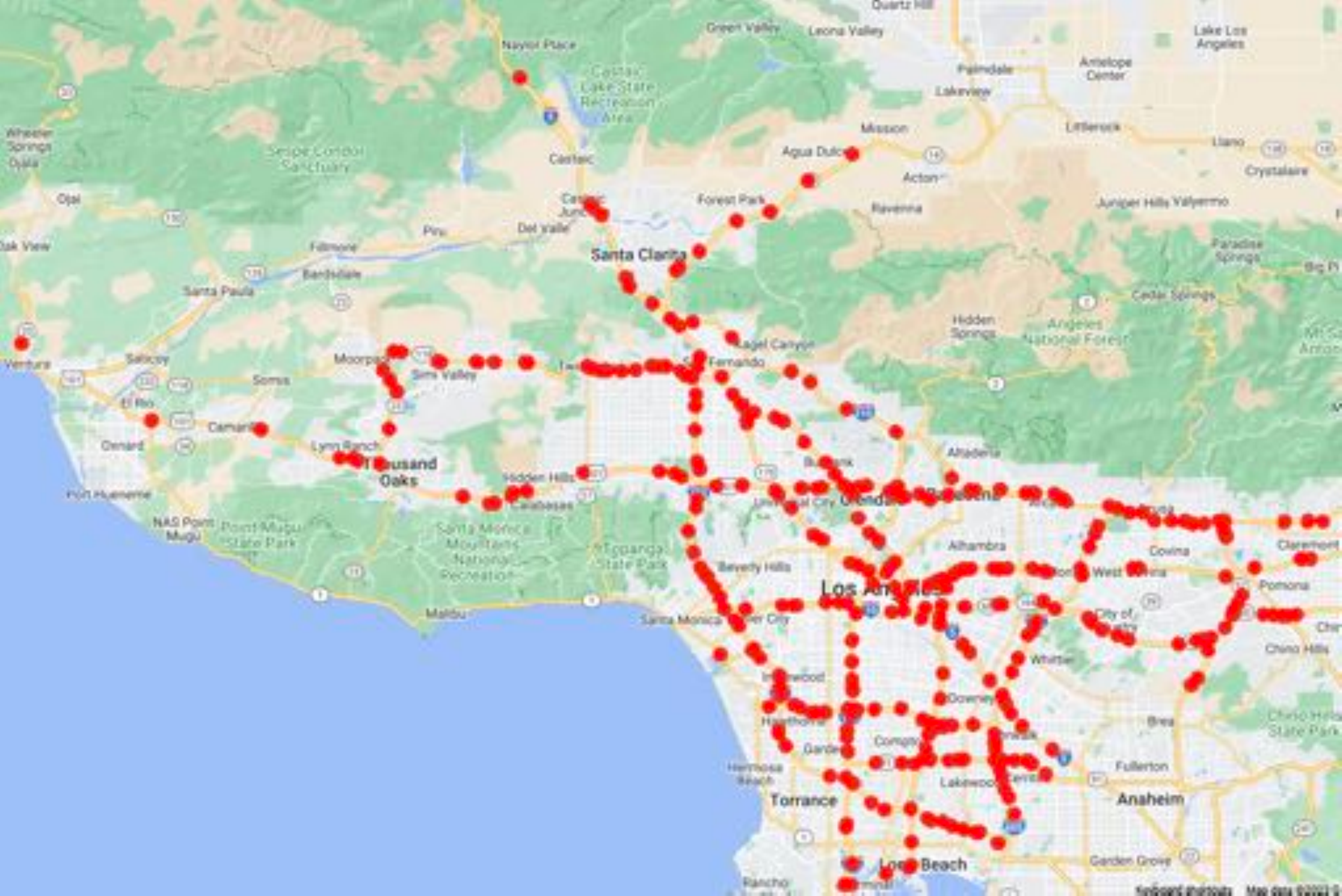}
         \caption{PEMS-07}
         \label{fig:sensor_all_07}
     \end{subfigure}
     \hspace{-1mm}
     \begin{subfigure}[b]{0.19\textwidth}
         \centering
         \includegraphics[width=\textwidth]{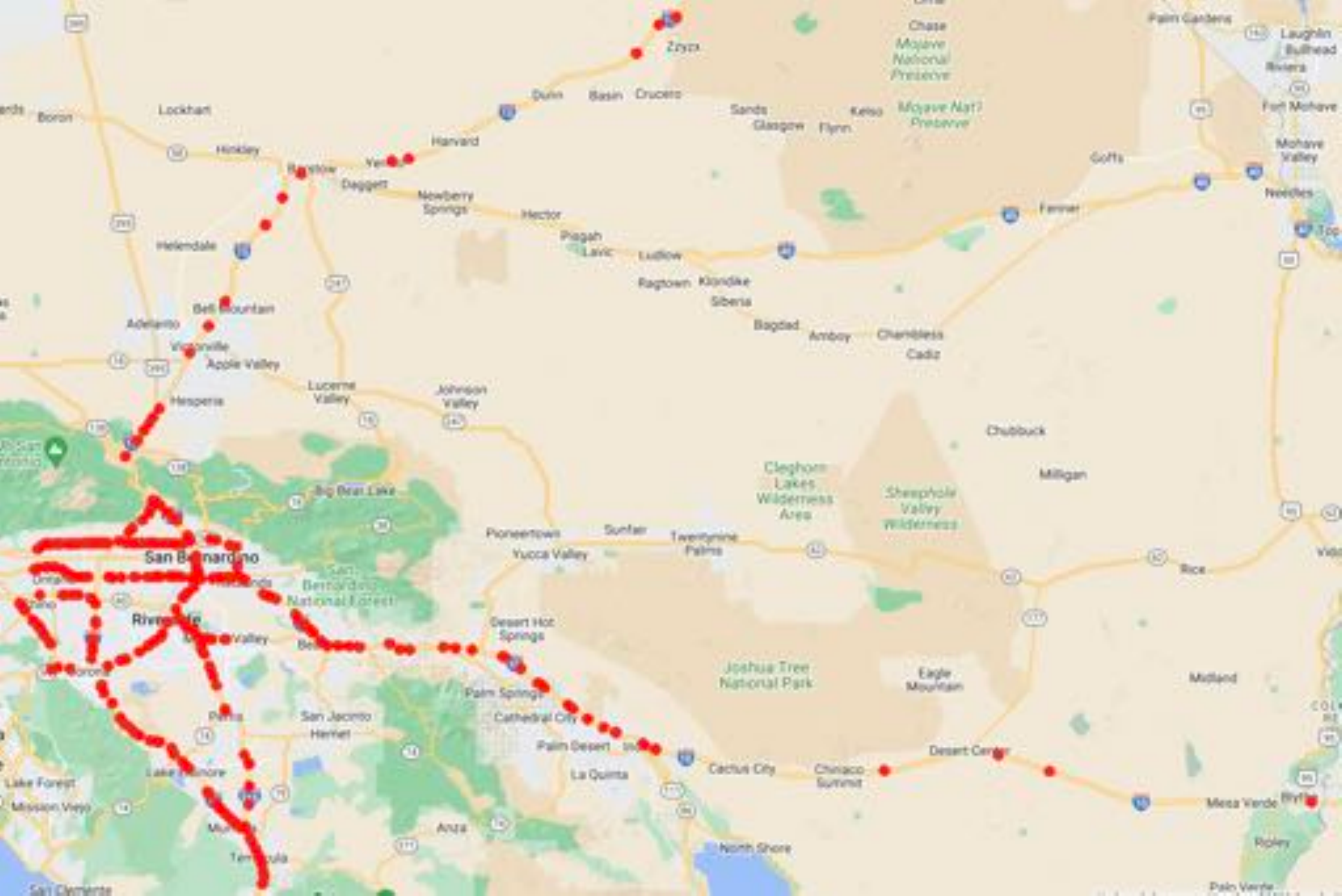}
         \caption{PEMS-08}
         \label{fig:sensor_all_08}
     \end{subfigure}
     \hspace{-1mm}
     \begin{subfigure}[b]{0.19\textwidth}
         \centering
         \includegraphics[width=\textwidth]{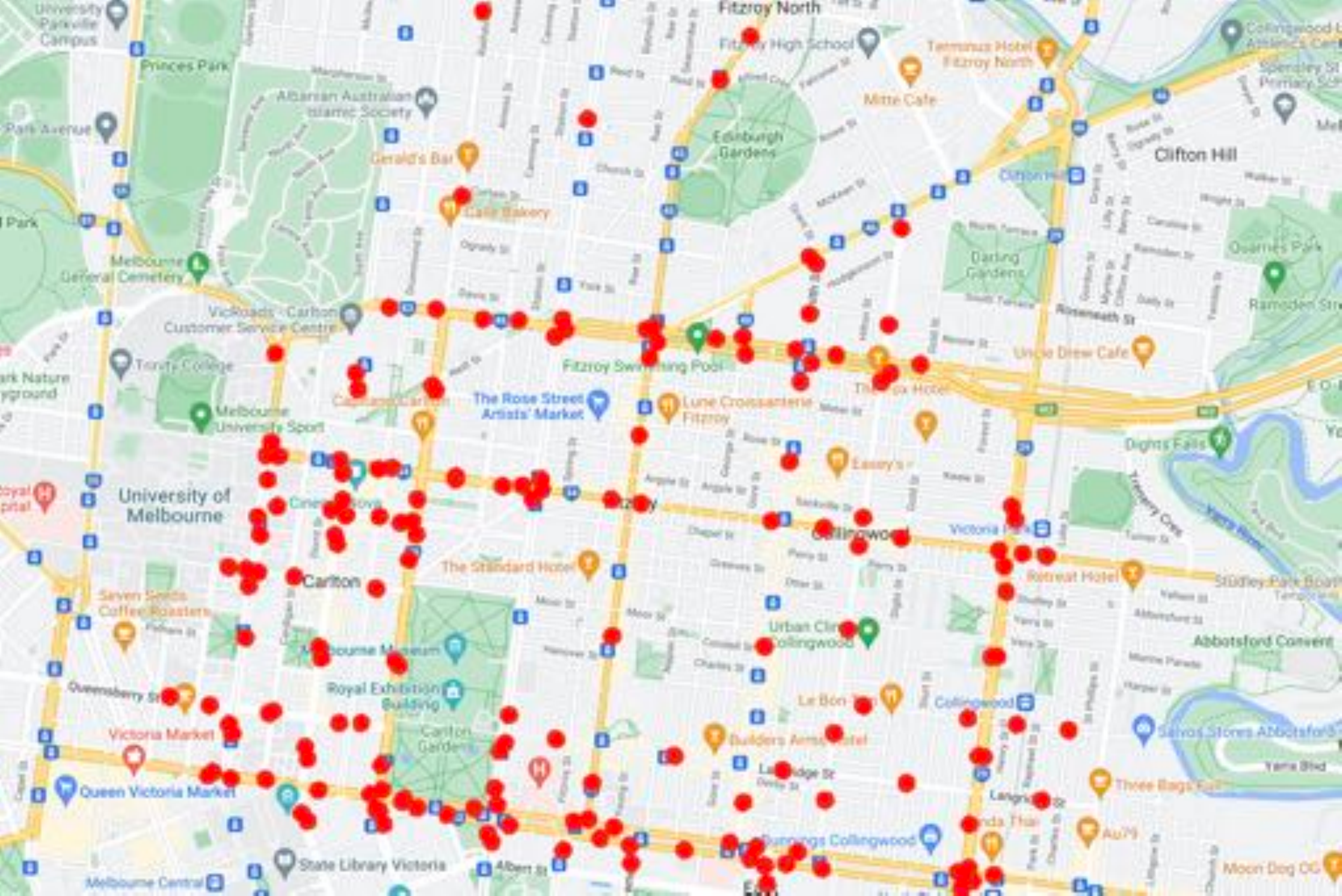}
         \caption{Melbourne}
         \label{fig:sensor_all_mel}
     \end{subfigure}
     \hspace{-1mm}
     \begin{subfigure}[b]{0.19\textwidth}
         \centering
         \includegraphics[width=\textwidth]{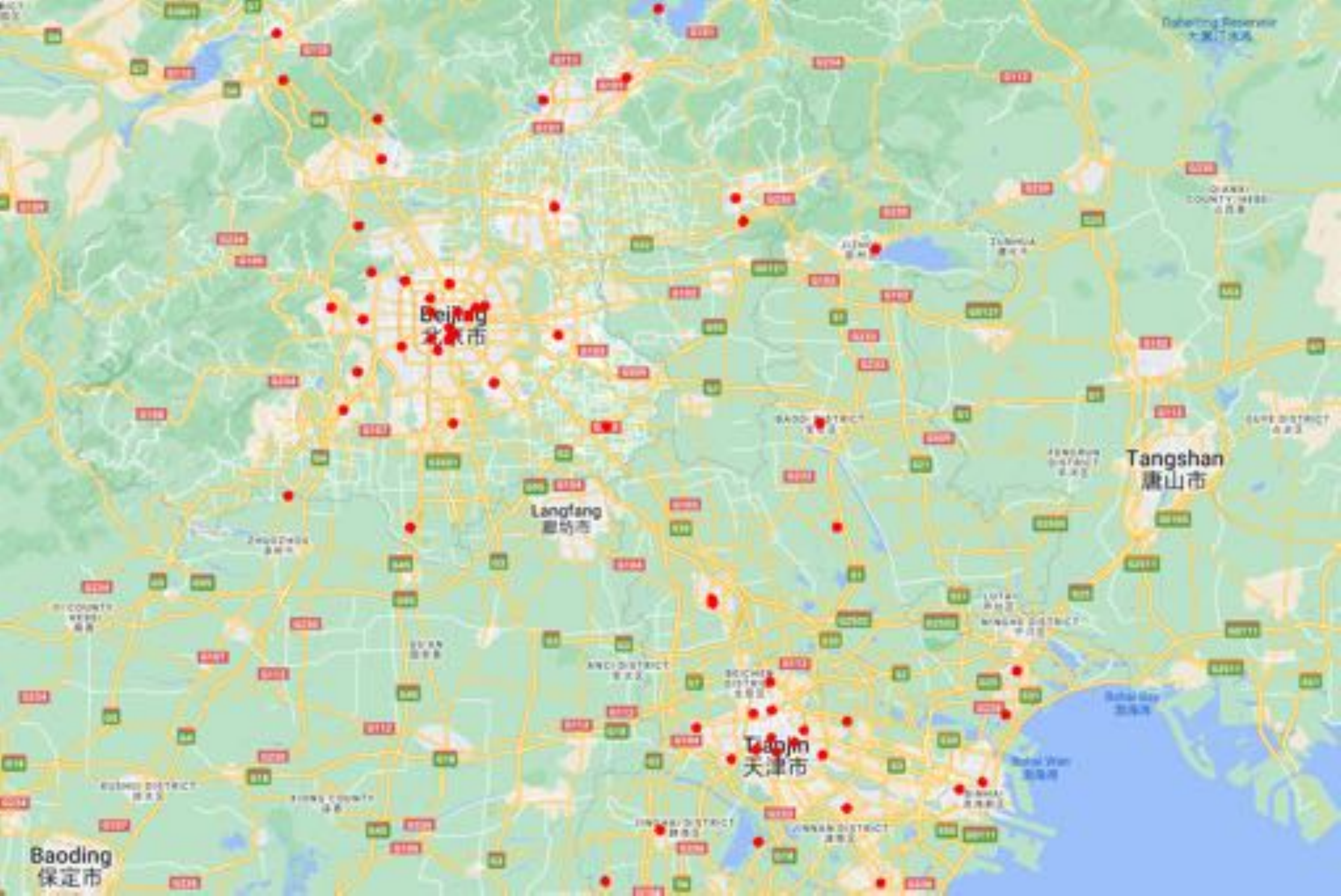}
         \caption{AirQ}
         \label{fig:sensor_all_air}
     \end{subfigure}
        \caption{Visualisations of sensor distribution}
        \label{fig:sensor_distribution}
\end{figure*}

We conduct experiments on three highway traffic datasets, an urban traffic dataset and an air quality dataset. 
\begin{itemize}
\item \textbf{PEMS-Bay}~\cite{dcrnn} contains traffic speed data collected from 325 sensors on highways in the Bay Area, California, between January and June 2017.  
\item \textbf{PEMS-07}~\cite{pems} contains traffic speed data collected by sensors on highways in Los Angeles. Following a previous study~\cite {gamcn}, we randomly sample 400 sensors and use their collected data between September and December 2022 as the dataset. 
\item \textbf{PEMS-08}~\cite{pems} contains traffic speed data collected by sensors on highways in the San Bernardino area, California. Similarly, we use data collected by 400 randomly sampled sensors between September and December 2022.
\item \textbf{Melbourne} contains traffic speed data collected by 182 sensors in Melbourne City, Australia, between July and September 2022, from the AIMES project~\cite{aimes}.
\item \textbf{AirQ}~\cite{air} contains pollutant concentration data (PM2.5) collected by 63 sensors in Beijing and Tianjin, which are two adjacent cities in China, between May 2014 and April 2015.
\end{itemize}
All traffic records collected from PEMS are given in 5-minute windows, i.e., 288 time slots per day, while the traffic records of the Melbourne dataset are given in 15-minute windows, i.e., 96 time slots per day. The air quality records are given in hourly windows, i.e., 24 time slots per day. Table~\ref{tab:dataset_info} summarises the dataset statistics and Fig.~\ref{fig:sensor_distribution} visualises the sensor distribution among all datasets. The region and road network information used for selective masking is obtained from OpenStreetMap~\cite{osm}. 

\begin{figure}[h!]
         \centering
  \includegraphics[width=\linewidth]{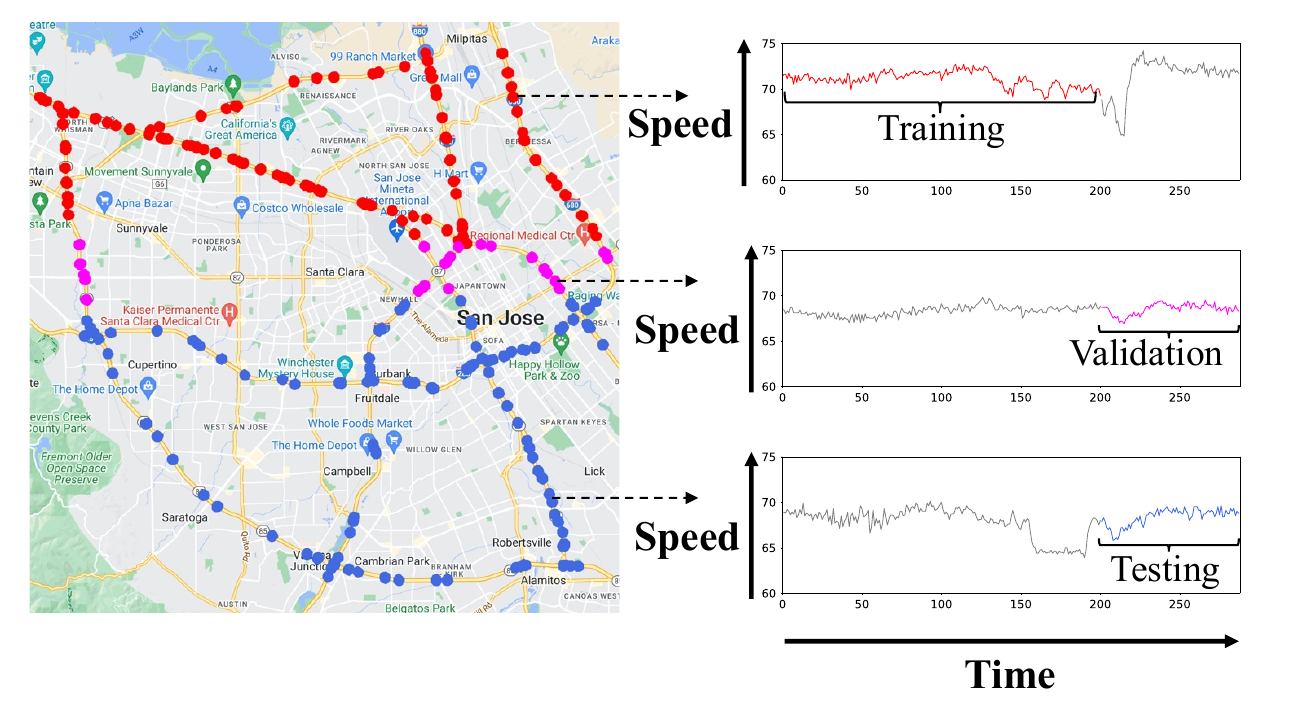}
         \caption{Data partitioning on PEMS-Bay from spatial (left; horizontal partitioning) and temporal (right) perspectives. The red, pink and blue dots on the map represent the observed locations for training, the observed locations for validation and the unobserved locations for testing, respectively.}
         \label{fig:train_valid_test}
\end{figure}

Following a baseline work~\cite{ignnk}, we use the records in the past two hours to make predictions for the next two hours, i.e., $T=T'=2 hours$ in Eq.~\ref{eq:task_definition} for the traffic datasets. Following another baseline work ~\cite{increase}, we use the records in the past 24 hours to make predictions for the next 24 hours, i.e., $T=T'=24 hours$ in Eq.~\ref{eq:task_definition} for the air quality dataset.
We split each dataset into three sets by 4:1:5 for training, validation and testing, where the locations within each set are adjacent to each other. Note that, the locations in the training set and the validation set are considered as observed locations, while those in the test set are considered as unobserved locations.
Dataset split is space-based, where the sensors are divided horizontally or vertically into three sets based on the sensor geo-coordinates. For each dataset, we create four different splits and report the average performance on each dataset.
We use the data recorded during the first 70\% of the time for the training, and the last 30\% of the time for testing.
Fig.~\ref{fig:train_valid_test} shows a dataset partitioning on PEMS-Bay and its temporal partitioning.

\begin{table}[!h]
\centering
  \caption{Dataset Statistics}
  \label{tab:dataset_info}
\begin{tabular}{p{1.5cm}|p{3cm}rr}
  \hlineB{3}
  Dataset & Time period & Interval & \#Sensors\\
  \hline \hline
  PEMS-Bay & 01/01/2017 - 30/06/2017 & 5 min & 325\\
  PEMS-07 & 01/09/2022 - 31/12/2022 & 5 min & 400\\
  PEMS-08 & 01/09/2022 - 31/12/2022 & 5 min & 400\\
  Melbourne & 01/07/2022 - 30/09/2022 & 15 min &182\\
  AirQ & 01/05/2014 - 30/04/2015 & 1 hr & 63\\
  \hlineB{3}
\end{tabular}
\end{table}

\subsubsection{Competitors.} There are no existing models for our proposed problem. We adopt the following adapted models for an empirical comparison with our proposed model \model: 
\begin{itemize}
\item \textbf{GE-GAN}~\cite{ge-gan} is a transductive data imputation method based on generative adversarial networks (GAN) that utilise a generator to generate estimating values and a discriminator to classify the real and generated values. 
\item \textbf{IGNNK}~\cite{ignnk} is an inductive graph neural network for spatial-temporal Kriging. 
\item \textbf{INCREASE}~\cite{increase} is an inductive graph representation learning network based on GRUs and the state-of-the-art for spatial-temporal Kriging. 
\end{itemize}

\subsubsection{Implementation Details.} We use the default settings of the baseline models from their source code. The baseline models were proposed for data imputation, while we aim for prediction. We change their ground truth to the future time window rather than the past time window to train the models and obtain the prediction. 

\begin{figure}[h!]
     \begin{subfigure}[b]{0.24\textwidth}
         \centering
         \includegraphics[width=\textwidth]{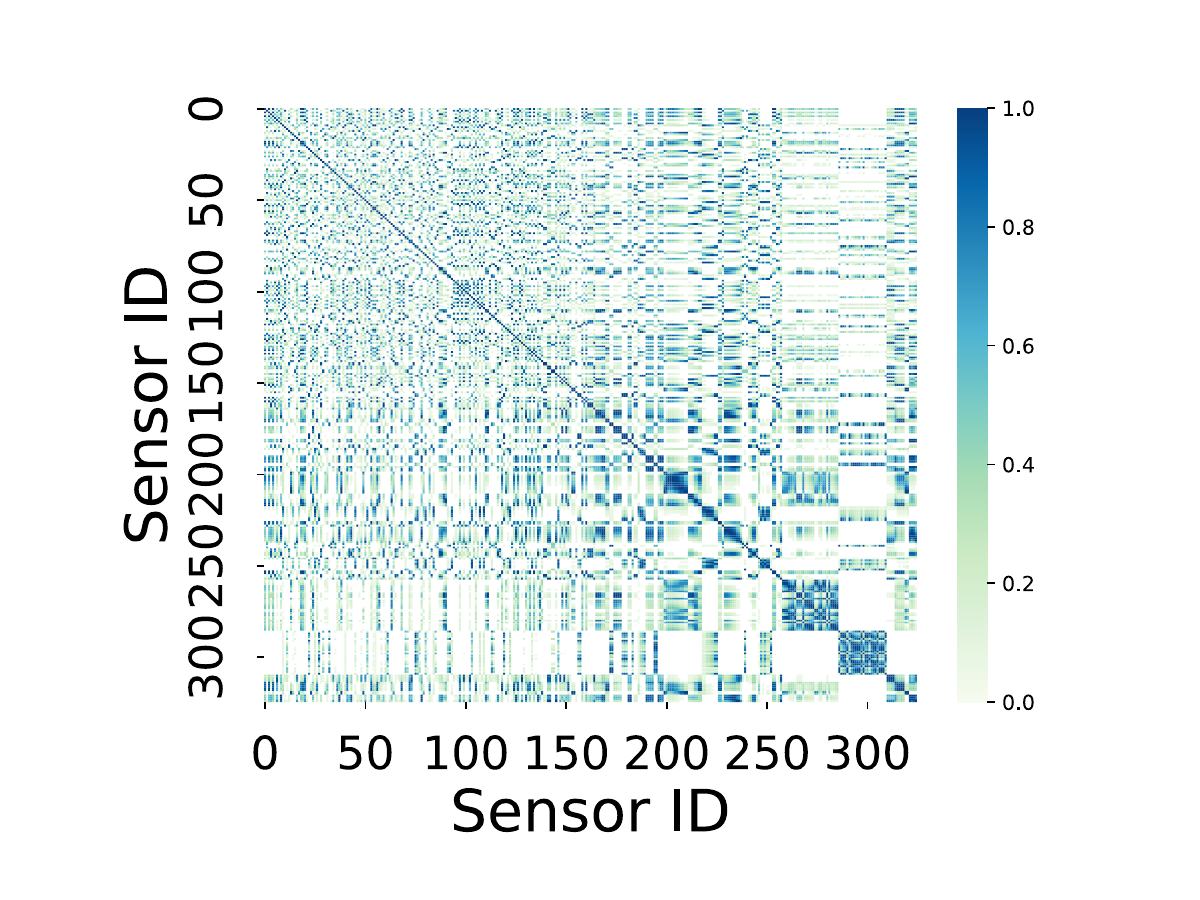}
         \caption{$A_s$}
         \label{fig:A_s}
     \end{subfigure}
     \hspace{-3mm}
     \begin{subfigure}[b]{0.24\textwidth}
         \centering
         \includegraphics[width=\textwidth]{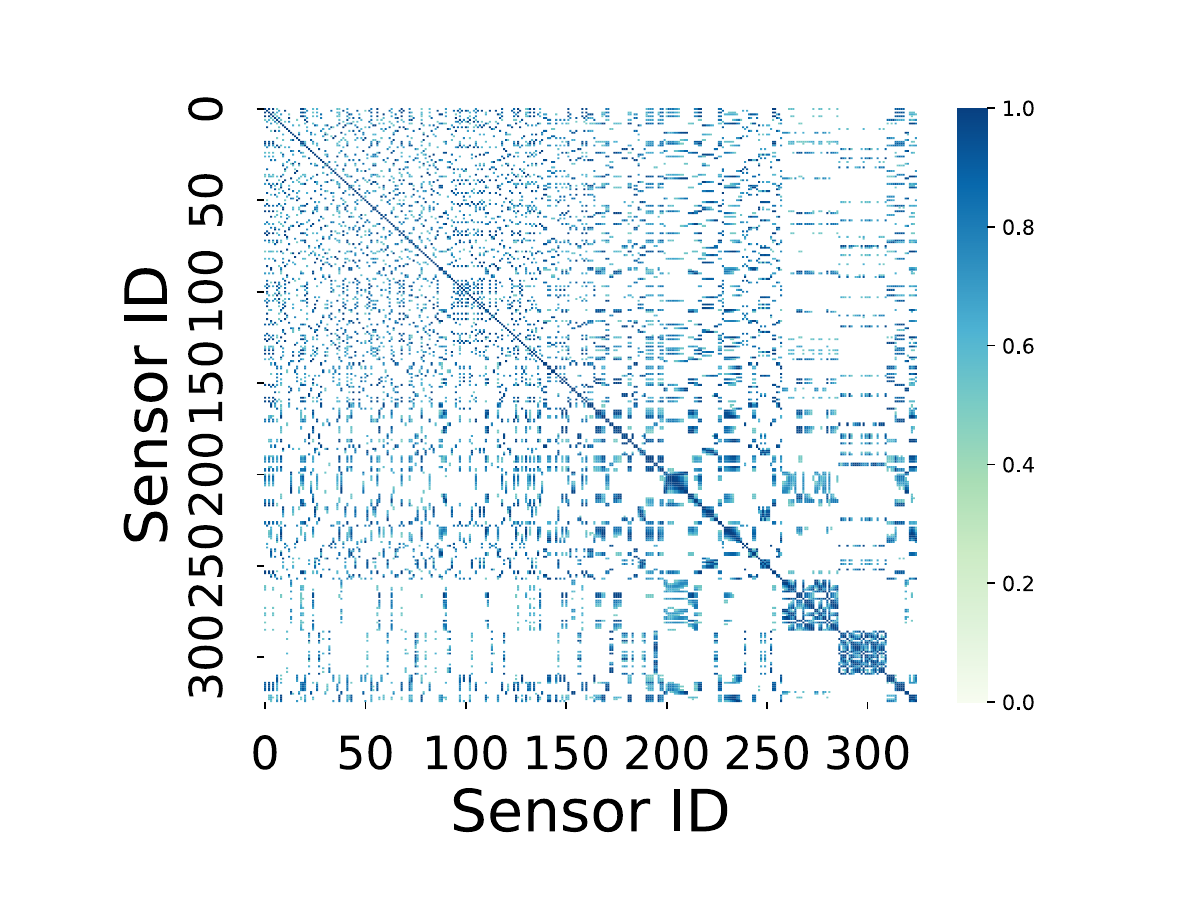}
         \caption{$A_{sg}$}
         \label{fig:A_sg}
     \end{subfigure}
        \caption{Visualization of adjacency matrices. The blank density in these figures reflects the sparsity of the adjacency matrix. The right figure has more blank space due to we use a larger threshold to limit sub-graph size.}
        \label{fig:visulization_matirx}
\end{figure}

We train our model using the Adam optimiser with the learning rate starting at 0.01. The batch size is 32. 
For the hyper-parameters in our models, $\tau$ is 0.5, $\sigma_m$ is 0.5, $\epsilon_{s}$ is 0.05 and $q_{kk}$ and $q_{ku}$ are set to 1. 
We leave the details of the other model hyper-parameters (i.e., $\lambda$, $\sigma_{sg}$, $r_{poi}$ and $K$) in Table~\ref{tab:param_setting}. These parameter values are obtained through grid search on the validation set, except $r_{poi}$ which is fine-tuned only based on the similarity between sub-graphs and the unobserved regions (i.e., $S^{sg}$). Also, parameter values can be shared among datasets with similar distributions, e.g., when only the number or the density of sensors is changed (cf., Table~\ref{tab:varying_sensors_num} and Table~\ref{tab:varying_sensors_density}).
In the experiments, we use different thresholds for the spatial-based matrices $A_s$ and $A_{sg}$. Fig.~\ref{fig:visulization_matirx} visualises the two adjacency matrices on PEMS-Bay.

The experiments are run on an NVIDIA Tesla V100 GPU.

\begin{table}[H]
\centering
  \caption{Parameter settings}
  \label{tab:param_setting}
  \resizebox{\columnwidth}{!}{
\begin{tabular}{l|ccccc}
\hlineB{3}
  Parameter & PEMS-Bay & PEMS-07 & PEMS-08 & Melbourne & AirQ\\
  \hline \hline
  $\lambda$ & 0.01 & 1 & 0.5 & 0.5 & 1\\
  \hline
  $\epsilon_{sg}$ & 0.5 & 0.7 & 0.5 & 0.4 & 0.6\\
  \hline
  $r_{poi}$ (m) & 200 & 500 & 500 & 50 & 500\\
  \hline
  $K$ & 35 & 35 & 35 & 45 & 5\\
\hlineB{3}
\end{tabular}
}
\end{table}

We adopt four commonly used metrics to evaluate model performance, including root mean squared error (RMSE), mean absolute error (MAE), mean absolute percentage error (MAPE) and R-square (R2). The first three measure the forecasting errors, while R2 measures how much better the model prediction results are compared with just using average observations as results~\cite{increase}. 

\subsection{Experimental Results}
\label{subsec:experiment_results}
We first compare the overall performance of our model with those of the baseline methods. Then, we report the results of an ablation study to verify the effectiveness of each module in \model. Finally, we study the impact of parameters to test model robustness. 

\subsubsection{Model Performance Comparison.} We first compare \model\ with the baseline methods.

\begin{table*}[!h]
\centering
  \caption{Overall model performance.  ``$\downarrow$'' (and ``$\uparrow$'') indicates that lower (and larger) values are better. The best baseline results are \underline{underlined}, and the best results of the proposed model \model\ are in \textbf{bold}.
  Improvement computes the errors reduced by the best variant of the proposed model compared with the best baseline model, where N/A indicates that the improvement results cannot be calculated, because of negative measurement values on the baseline methods.
  }
\label{tab:overallperformance}
\resizebox{\textwidth}{!}{
\begin{tabular}
{l|c|ccc|cccc|r}
  \hlineB{3}
  Dataset & Metric & GE-GAN & IGNNK & INCREASE & \model-RNC &\model-NC & \model-R & \textbf{\model} & 	Improvement\\
  \hline \hline
  \multirow{4}{*}{PEMS-Bay} &  RMSE$\downarrow$ & 31.184 & 9.611 & \underline{8.820} & 8.626 & 8.642 & 8.628 & \textbf{8.610} & +2.38\%\\
   & MAE$\downarrow$ & 26.020 & 5.616 & \underline{5.243} & 5.352 & 5.352 & \textbf{5.120} & 5.237 & +0.11\%\\
  & MAPE$\downarrow$ & 0.432 & 0.145 & \underline{0.133} & 0.130 & 0.131 & \textbf{0.128} & 0.130 & +2.26\%\\
  & R2$\uparrow$ & -9.042 & 0.063 & \underline{0.195} & 0.228 & 0.226 & 0.230 & \textbf{0.231} & +18.46\%\\
  \hline
  \multirow{4}{*}{PEMS-07} & RMSE$\downarrow$ & 21.103 & 11.556 & \underline{8.465} & 8.516 & 8.512 & 8.444 & \textbf{8.390} & +0.89\%\\
  & MAE$\downarrow$ & 15.647 & 9.167 & \underline{5.495} & 5.352 & 5.411 & 5.417 & \textbf{5.111} & +6.99\%\\
  & MAPE$\downarrow$ & 0.273 & 0.181 & \underline{0.125} & 0.125 & 0.126 & 0.125 & \textbf{0.123} & +1.60\%\\
  & R2$\uparrow$ & -4.351 & -0.715 & \underline{0.155} & 0.156 & 0.142 & 0.157 & \textbf{0.169} & +9.03\%\\
  \hline
  \multirow{4}{*}{PEMS-08} &  RMSE$\downarrow$ & 23.409 & 10.599 & \underline{8.275} & 8.021 & 7.937 & 8.018 & \textbf{7.925} & +4.23\%\\
  & MAE$\downarrow$ & 17.611 & 7.934 & \underline{5.019} & 4.982 & 4.962 &5.006 & \textbf{4.899} & +2.39\%\\
  & MAPE$\downarrow$ & 0.299 & 0.158 & \underline{0.116} & 0.115 & 0.115 & 0.116 & \textbf{0.114} & +1.72\%\\
  & R2$\uparrow$ & -6.531 & -0.678 & \underline{0.056} & 0.114 & 0.134 & 0.115 & \textbf{0.136} & +142.86\%\\
  \hline
  \multirow{4}{*}{Melbourne} & RMSE$\downarrow$ & \underline{10.064} & 14.635 & 10.321 & 9.844 & 9.884 & 10.07 & \textbf{9.175} & +8.83\%\\
 & MAE $\downarrow$ & \underline{7.780} & 12.511 & 8.302 & 7.803 & \textbf{7.188} & 7.882 & 7.308 & +7.61\%\\
  & MAPE $\downarrow$ & \underline{0.369} & 0.746 & 0.453 & 0.399 & \textbf{0.366} & 0.389 & 0.388 & +0.81\%\\
 & R2$\uparrow$ & \underline{-0.175} & -2.355 & -0.266 & -0.120 & \textbf{0.027} & -0.171 & \textbf{0.027} & N/A\\

\hline
  \multirow{4}{*}{AirQ} & RMSE$\downarrow$ & 295.579 & 74.873 &\underline{73.977} & 69.956 & 68.126  & 68.968 & \textbf{67.571} & +8.66\%\\
 & MAE $\downarrow$ & 244.824 & \underline{52.726} & 56.165 & 50.301 & 48.451 & 49.302 & \textbf{48.141} & +14.29\% \\
  & MAPE $\downarrow$ & 9.142 & \underline{1.459} & 2.168 & 1.789 & \textbf{1.643} & 1.737 & 1.692 & -12.61\%\\
 & R2$\uparrow$ & -17.917 & -0.067 & \underline{-0.024} & 0.074 & 0.123 & 0.099  & \textbf{0.141} & N/A\\
  \hlineB{3}
\end{tabular}
}
\end{table*}

(1)~\emph{Overall Results.}: Table~\ref{tab:overallperformance} summarises the overall performance results. \model~ and its variants including the base model \model-RNC (detailed in Section~\ref{subsubsec:ablation_study}) outperform all the competitors on all four datasets, except for the MAPE measure on AirQ. 

GE-GAN is a transductive method that generates values for unobserved locations utilising their similar locations based on graph embeddings. It is difficult to find similar locations when there are many unobserved locations in a large area, resulting in poor forecasting accuracy. In urban areas such as Melbourne City, GE-GAN outperforms the other two baseline models because this area is relatively small.

IGNNK is an inductive model with GNNs to model spatial correlations and 1-D convolution neural networks to capture temporal correlations. It struggles in our task because data missing at continuous locations makes it difficult for the GNNs to learn the spatial correlation patterns. Although it has a slightly low MAPE on AirQ, its MAE and RMSE are still much larger than those of our model \model\ on this dataset. This can be explained by that a lower MAE happens on smaller observations while a higher MAE happens on larger observations.

INCREASE, the state-of-the-art spatial-temporal Kriging model, learns heterogeneous spatial relations and diverse temporal patterns, which presents the best performance among the baseline models. However, it is still outperformed by our model \model. Our model reduces forecasting errors by up to 14\% on the AirQ dataset and increases R2 by up to 142\% on the PEMS-08 dataset because of our temporal adjacency matrix to model the temporal similarity, our selective masking module to model spatial and semantic similarity, and contrastive learning to enhance model robustness.

We also report the model training and testing time over the traffic datasets. We omit the running time results on AirQ due to its small scale. 
The training time of all models is at the same scale. GE-GAN requires more training epochs to converge. However, when it comes to testing time, GE-GAN and \model\ are faster than IGNNK and INCREASE. 

\begin{table}[h!]
\centering
  \caption{Model training time}
  \label{tab:running_time}
  \resizebox{\columnwidth}{!}{
  \begin{tabular}{llcccc}
\hlineB{3}
  Model & Time & PEMS-Bay & PEMS-07 & PEMS-08 & Mel.\\
  \hline \hline
  \multirow{2}{*}{GE-GAN} & Train (h) & 4.4 & 4.1 & 4.1 & 0.3\\
  & Test (s) & 0.9 & 0.7 & 0.8 & 0.1\\
  \hline
  \multirow{2}{*}{IGNNK} & Train (h) & 0.3 & 0.2 & 0.2 & 0.1\\
  & Test (s) & 8.3 & 7.8 & 8.8 & 2.4 \\
  \hline
  \multirow{2}{*}{INCREASE} & Train (h) & 0.3 & 0.2 & 0.2 & 0.2\\
  & Test (s) & 9.5 & 7.3 & 7.5 & 4.0 \\
  \hline
  \multirow{2}{*}{\model} & Train (h) & 1.1 & 1.9 & 2.2 & 0.3\\
  & Test (s) & 1.6 & 1.3 & 1.2 & 0.3\\
\hlineB{3}
\end{tabular}
}
\end{table}

\begin{figure}[h!]
     \centering
     \begin{subfigure}[b]{0.23\textwidth}
         \centering
         \includegraphics[width=\textwidth]{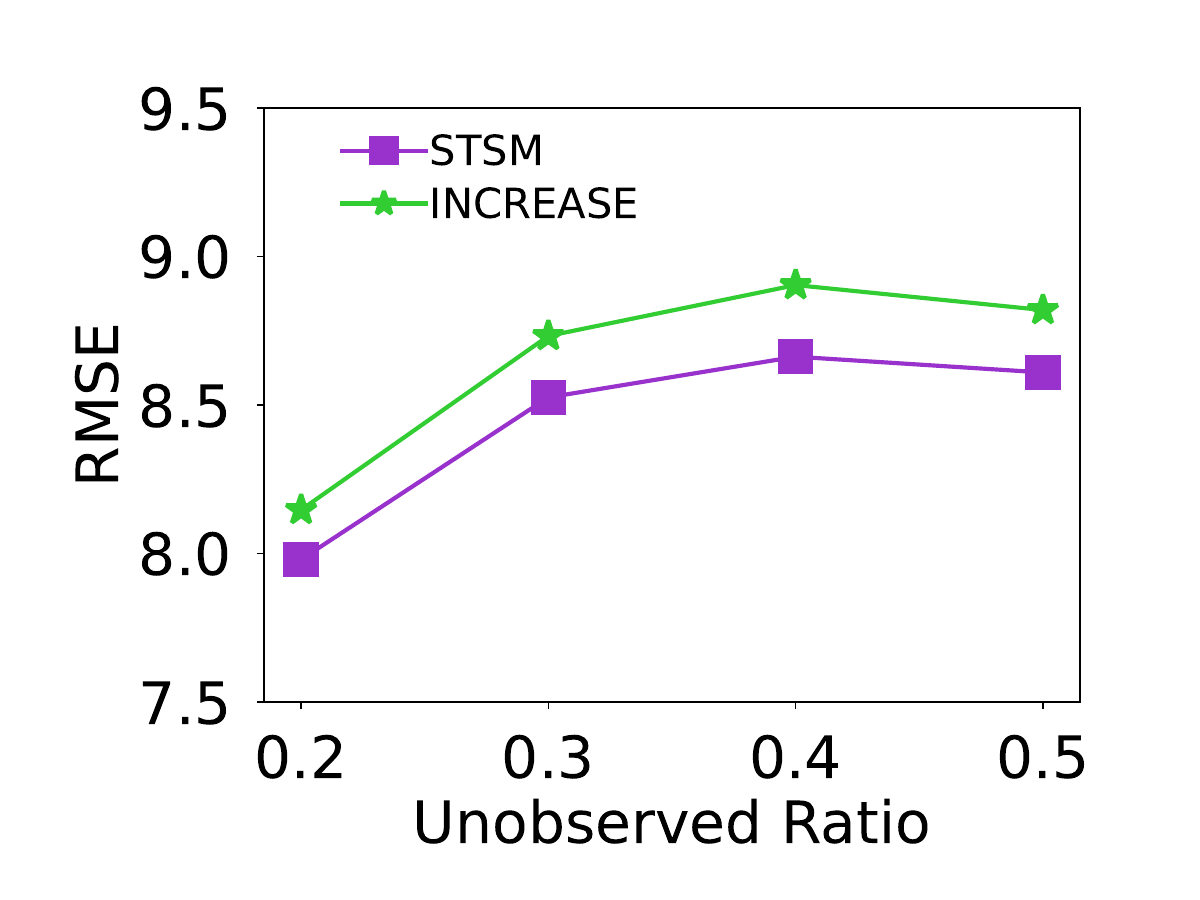}
         \hspace{-2mm}
         \caption{PEMS-Bay}
         \label{fig:unknow_r_bay}
     \end{subfigure}
     \begin{subfigure}[b]{0.23\textwidth}
         \centering
         \includegraphics[width=\textwidth]{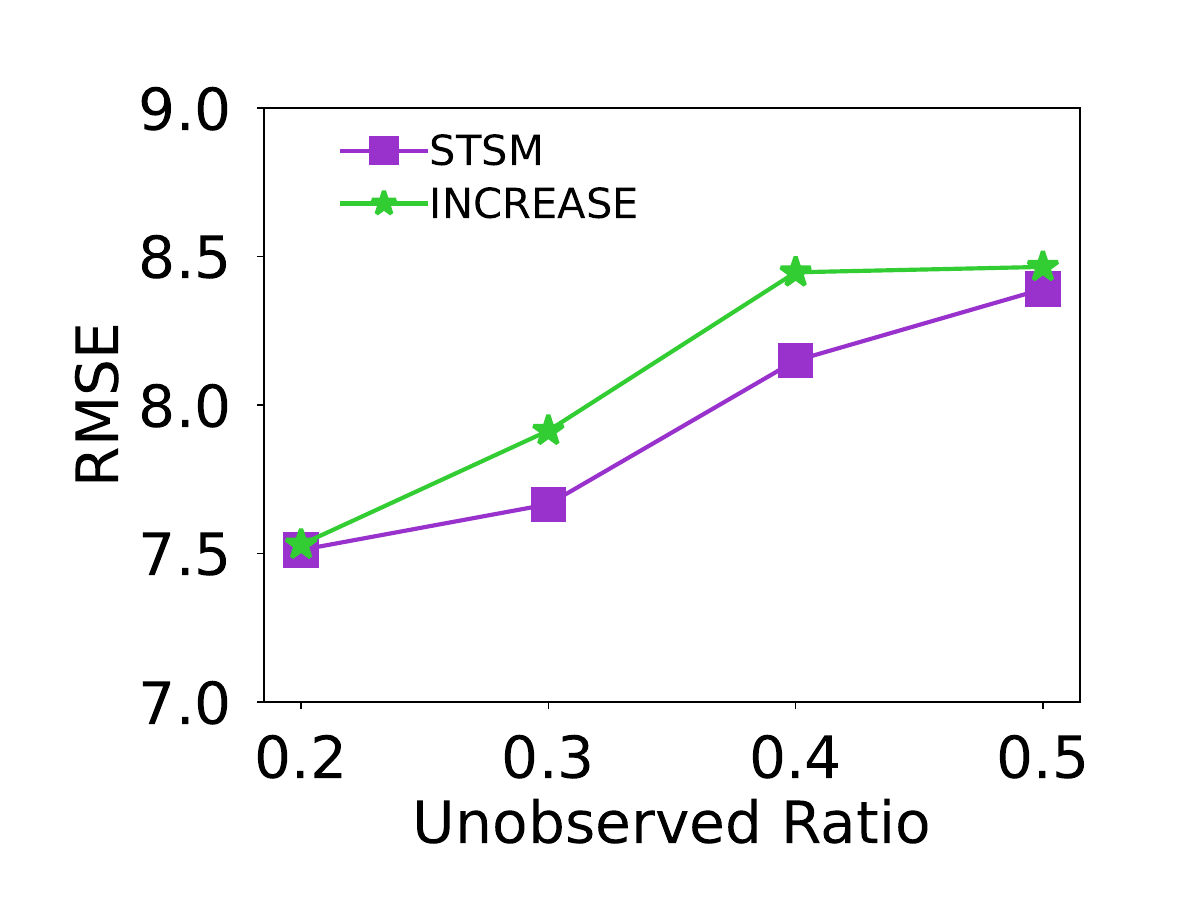}
         \hspace{-2mm}
         \caption{PEMS-07}
         \label{fig:unknow_r_07}
     \end{subfigure}
     \begin{subfigure}[b]{0.23\textwidth}
         \centering
         \includegraphics[width=\textwidth]{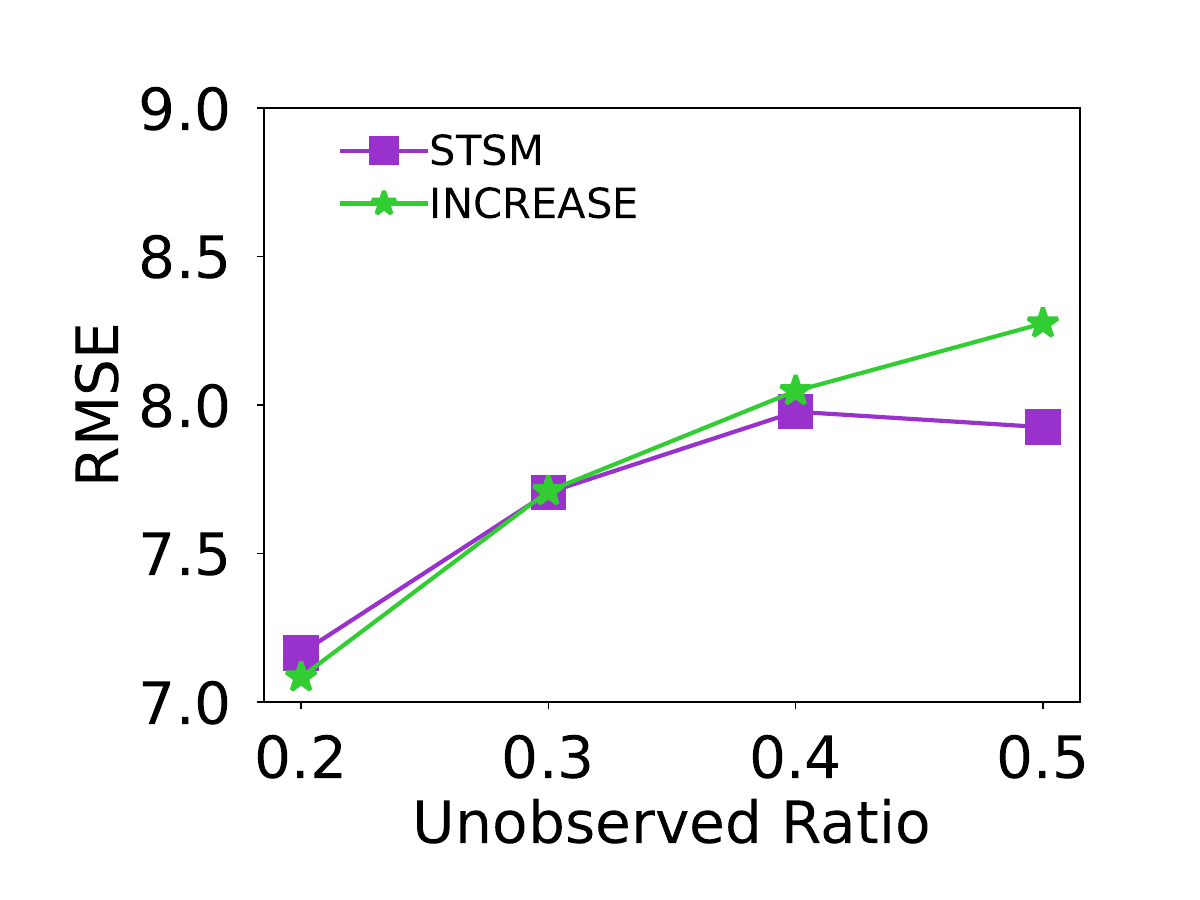}
         \hspace{-2mm}
         \caption{PEMS-08}
         \label{fig:unknow_r_08}
     \end{subfigure}
     \begin{subfigure}[b]{0.24\textwidth}
         \centering
         \includegraphics[width=\textwidth]{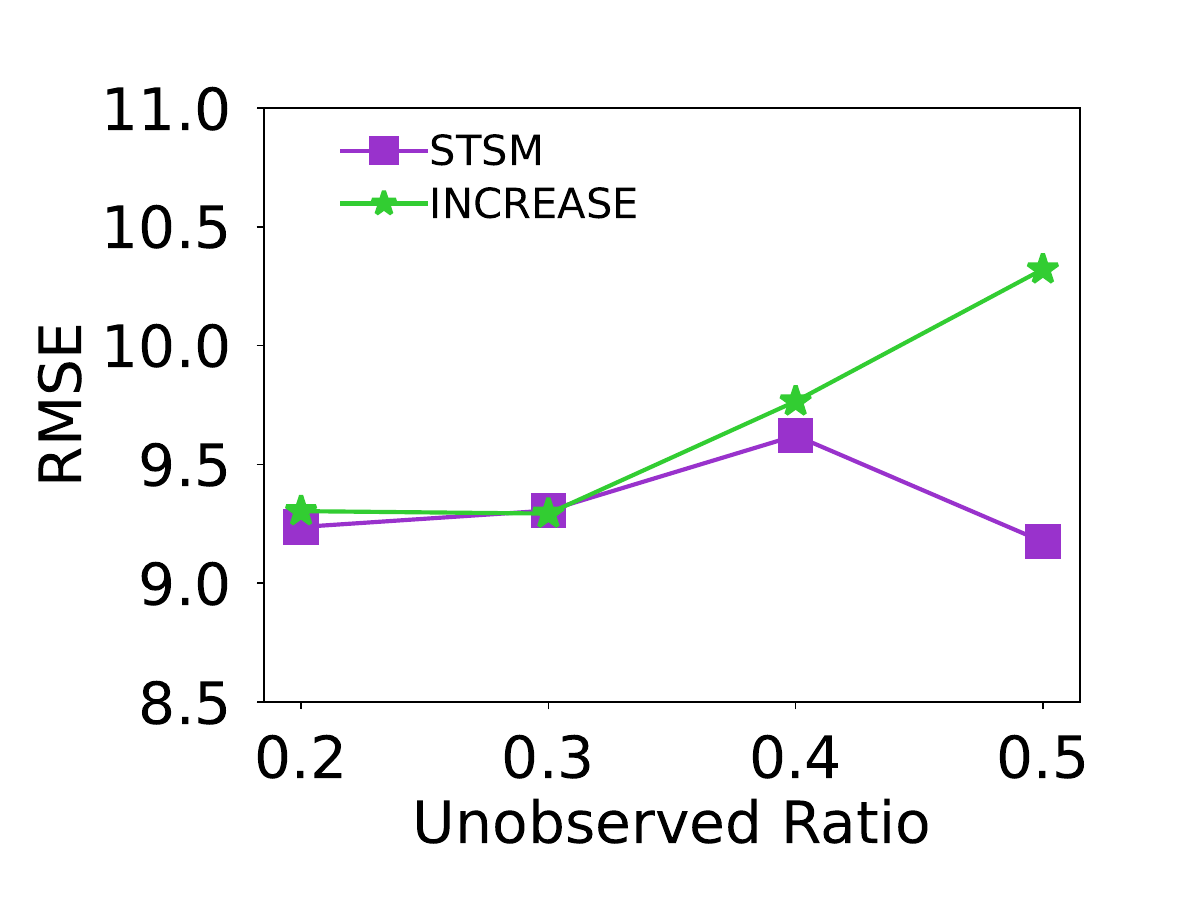}
         \hspace{-2mm}
         \caption{Melbourne}
         \label{fig:unknow_r_mel}
     \end{subfigure}
     \begin{subfigure}[b]{0.23\textwidth}
         \centering
         \includegraphics[width=\textwidth]{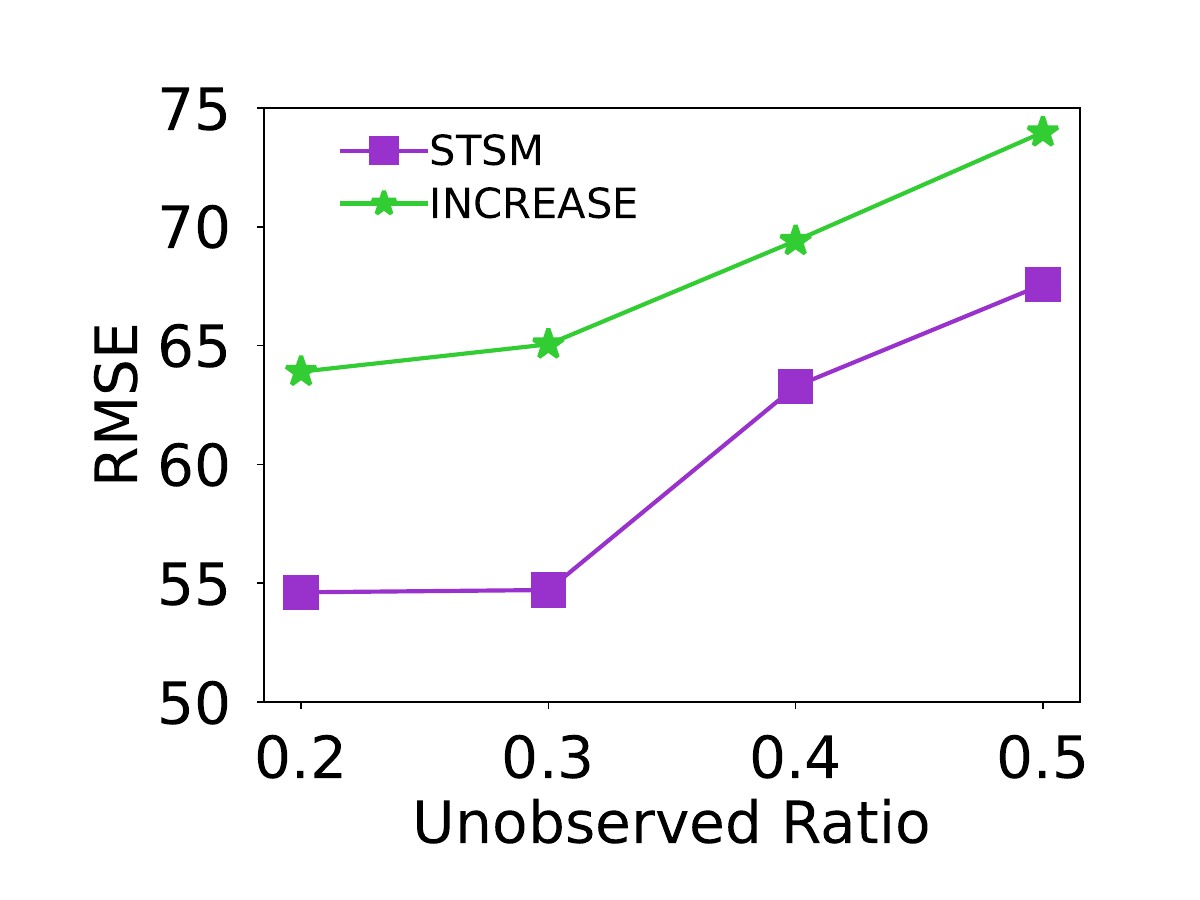}
         \hspace{-2mm}
         \caption{AirQ}
         \label{fig:unknow_r_air}
     \end{subfigure}
        \caption{Model performance vs. unobserved ratio}
        \label{fig:Unknwo_Ratio}
\end{figure}

(2)~\emph{Varying the unobserved ratio.} We vary the unobserved ratio from 0.2 to 0.5 on all datasets, i.e., from 20\% to 50\% of all the sensor locations in a dataset are treated as unobserved locations. Same as before, we split each dataset horizontally or vertically and report the average performance on the four setups (each split creates two alternative settings of training and testing sets). Fig.~\ref{fig:Unknwo_Ratio} presents the results. Since INCREASE has the best performance among the baselines when varying the unobserved ratio, we only show its results for this set of experiments. \model\ outperforms INCREASE in all settings, except when 20\% of the locations are considered unobserved on PEMS-08. We notice that sometimes the prediction errors drop even with a higher ratio of unobserved locations. This is because some unobserved locations are easier than others to predict. Including such unobserved locations reduces the mean prediction error. Here, we have only shown results in RMSE. Results on other metrics show similar patterns, which are omitted for conciseness. The same applies to the experiments below. 

(3)~\emph{Varying the number of sensors.} We merge PEMS08 and PEMS07 into a larger region, such that we can vary the number of sensors from 200 to 800 by vertically splitting the space (and hence the sensor locations) into four equal-sized partitions (i.e., each partition contains 200 sensors) based on geo-coordinates. 
Table~\ref{tab:varying_sensors_num} shows reports how the model prediction errors change with more sensors are added into the dataset. We see that our model \model\ consistently outperforms all three baseline models in terms of RMSE and R2. 

\begin{table}[!h]
\centering
  \caption{Varying the number of sensors.  ``$\downarrow$'' (and ``$\uparrow$'') indicates that lower (and larger) values are better. The best baseline results are \underline{underlined}, and the best model results are in \textbf{bold}.}
\label{tab:varying_sensors_num}
\resizebox{\columnwidth}{!}{
\begin{tabular}
{c|l|rrrr}
  \hlineB{3}
  \#Sensor number & Model & RMSE$\downarrow$ & MAE$\downarrow$ & MAPE$\downarrow$ & R2$\uparrow$\\
  \hline \hline
  \multirow{4}{*}{200} & GE-GAN & 24.442 & 18.348 & 0.309 & -7.828\\
   & IGNNK & 10.845 & 7.774 & 0.157 & -0.950\\
  & INCREASE  & \underline{7.800} & \underline{4.816} & \underline{0.110} & \underline{0.129}\\
  & \model & \textbf{7.775} & \textbf{4.649} & \textbf{0.109} & \textbf{0.133}\\
  \hline
  \multirow{4}{*}{400} & GE-GAN & 21.929 & 16.244 & 0.283 & -4.702 \\
   & IGNNK & 15.896 & 12.709 & 0.238 & -2.998\\
  & INCREASE & \underline{8.880} & \textbf{\underline{5.115}} & \textbf{\underline{0.130}} & \underline{0.115}\\
  & \model & \textbf{8.718} & 5.384 & 0.131 & \textbf{0.141}\\
  \hline
  \multirow{4}{*}{600} & GE-GAN &23.105 & 17.043 & 0.296 & -5.243 \\
   & IGNNK & 11.492 & 9.428 & 0.229 & -1.089\\
  & INCREASE & \underline{8.658} & \textbf{\underline{5.494}} & \textbf{\underline{0.129}} & \underline{0.142}\\
  & \model & \textbf{8.629} & 5.548 & 0.130 & \textbf{0.148}\\
  \hline
  \multirow{4}{*}{800} & GE-GAN & 21.801 & 15.867 & 0.280 & -4.349\\
   & IGNNK & 12.113 & 8.530 & 0.172 & -1.295\\
  & INCREASE & \underline{8.370} & \underline{5.055} & \underline{0.119} & \underline{0.113} \\
  & \model & \textbf{8.134} & \textbf{5.008 }& \textbf{0.118} & \textbf{0.165}\\
  \hlineB{3}
\end{tabular}
}
\end{table}

(4)~\emph{Varying the density of sensors.} We further vary the number of sensors from 200 to 964 (which is the maximum number of sensors) on PEMS-08, to test the impact of the density of the sensors.  
The results in Table~\ref{tab:varying_sensors_density} show that \model\ again outperforms all baseline models in almost all cases (i.e., 19 out of 20), further confirming the robustness of the model.

\begin{table}[!h]
\centering
  \caption{Varying the density of sensors.  ``$\downarrow$'' (and ``$\uparrow$'') indicates that lower (and larger) values are better. The best baseline results are \underline{underlined}, and the best model results are in \textbf{bold}.}
\label{tab:varying_sensors_density}
\resizebox{\columnwidth}{!}{
\begin{tabular}
{c|l|rrrr}
  \hlineB{3}
  \#Sensor number & Model & RMSE$\downarrow$ & MAE$\downarrow$ & MAPE$\downarrow$ & R2$\uparrow$\\
  \hline \hline
  \multirow{4}{*}{200} & GE-GAN & 24.293 & 18.102 &0.306 & -7.201\\
   & IGNNK & 13.847 & 9.981 & 0.194 & -2.623\\
  & INCREASE  & \underline{8.391} & \underline{5.188} & \underline{0.117} & \underline{0.016}\\
  & \model & \textbf{7.889} & \textbf{4.707} & \textbf{0.111} & \textbf{0.134}\\
  \hline
  \multirow{4}{*}{400} & GE-GAN & 23.409 & 17.611 & 0.299 & -6.531 \\
   & IGNNK & 10.599 & 7.934 & 0.158 & -0.678\\
  & INCREASE & \underline{8.275} & \underline{5.019} & \underline{0.116} & \underline{0.056}\\
  & \model & \textbf{7.925} & \textbf{4.899} & \textbf{0.114} & \textbf{0.136}\\
  \hline
  \multirow{4}{*}{600} & GE-GAN & 23.933 & 17.695 & 0.298 & -7.400\\
   & IGNNK & 12.348 & 9.198 & 0.175 & -1.745\\
  & INCREASE & \underline{7.982} & \underline{5.109} & \underline{0.110} & \underline{0.061}\\
  & \model & \textbf{7.708 }& \textbf{4.936} & \textbf{0.110} & \textbf{0.127} \\
  \hline
  \multirow{4}{*}{800} & GE-GAN & 22.711 & 17.016 & 0.288 & -6.405\\
   & IGNNK & 13.172 & 10.203 & 0.191 & -2.006\\
  & INCREASE & \underline{8.055} & \textbf{\underline{4.847}} & \underline{0.114} & \underline{0.072}\\
  & \model & \textbf{7.841} & 4.942 & \textbf{0.114} & \textbf{0.119}\\
  \hline
  \multirow{4}{*}{964} & GE-GAN & 22.180 & 16.685 & 0.283 &-6.186 \\
   & IGNNK & 12.476 & 9.735 & 0.184 & -1.639\\
  & INCREASE & \underline{8.052} & \underline{4.851} & \underline{0.114} & \underline{0.064}\\
  & \model & \textbf{7.831} & \textbf{4.757} & \textbf{0.111} & \textbf{0.113}\\
  \hlineB{3}
\end{tabular}
}
\end{table}

\subsubsection{Ablation Study} 
\label{subsubsec:ablation_study}
We conduct an ablation study with three variants of \model:

\textbf{\model-NC} disables the contrastive learning module. 

\textbf{\model-R} replaces the selective masking module with a random masking module that randomly chooses a location and its 1-hop neighbours to be masked until the target masking ratio is reached. 

\textbf{\model-RNC} (our base model as described in Section~\ref{sec:Proposed basic model}) replaces the selective masking module with the random masking module and disables contrastive learning. 

(1)~\emph{Impact of selective masking.} As Table~\ref{tab:overallperformance} shows, \model~outperforms \model-R among all datasets, except for the MAPE and MAE on PEMS-Bay. We further compare the similarity between the masked sub-graphs and the unobserved region in the training process. Table~\ref{tab:sim_gain} presents the results, which shows that selective masking can guide the model to mask sub-graphs with higher similarities to the unobserved regions. 

We also compare the performance of \model-NC and \model-RNC. \model-NC outperforms \model-RNC on PEMS-08, Melbourne and AirQ. On PEMS-Bay and PEMS-07, \model-NC and \model-RNC yield similar performance. 
These results confirm the importance of the selective masking module.

(2)~\emph{Impact of contrastive learning.} Table~\ref{tab:overallperformance} shows that \model\ outperforms \model-NC over all freeway traffic datasets. On the urban datasets (i.e., Melbourne and AirQ), \model\ has better RMSE while \model-NC is better at MAE or MAPE. 
Besides, \model-R outperforms \model-RNC in most cases (14 out of 20).
These results confirm that contrastive learning is also an important component that contributes to the strong model performance.

\begin{table}[h!]
\centering
  \caption{Similarity gain compared with random masking}
  \label{tab:sim_gain}
\resizebox{\columnwidth}{!}{
\begin{tabular}{c|ccccc}
  \hlineB{3}
  Dataset  & PEMS-Bay & PEMS-07 & PEMS-08 & Mel. & AirQ\\
  \hline \hline
  Sim. Gain (\%) & 9.35 & 14.76 & 5.39 & 6.87 & 19.66\\
  \hlineB{3}
\end{tabular}
}
\end{table}

\subsubsection{Parameter Study}
We test the impact of $K$ and $\epsilon_{sg}$.

\begin{figure}[!h]
     \centering
     \begin{subfigure}[b]{0.24\textwidth}
         \centering
         \includegraphics[width=\textwidth]{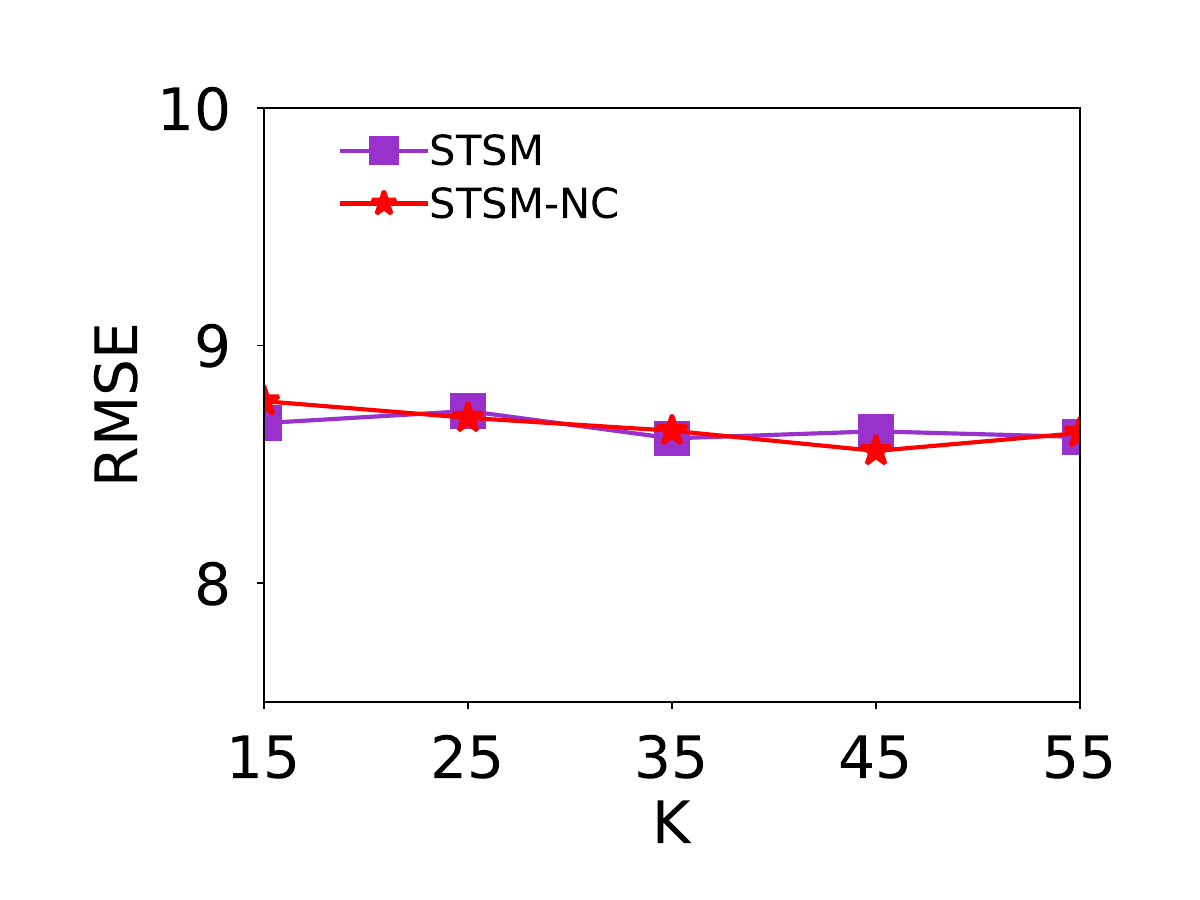}
         \hspace{-2mm}
         \caption{PEMS-Bay}
         \label{fig:K_bay}
     \end{subfigure}
     \hspace{-3mm}
     \begin{subfigure}[b]{0.24\textwidth}
         \centering
         \includegraphics[width=\textwidth]{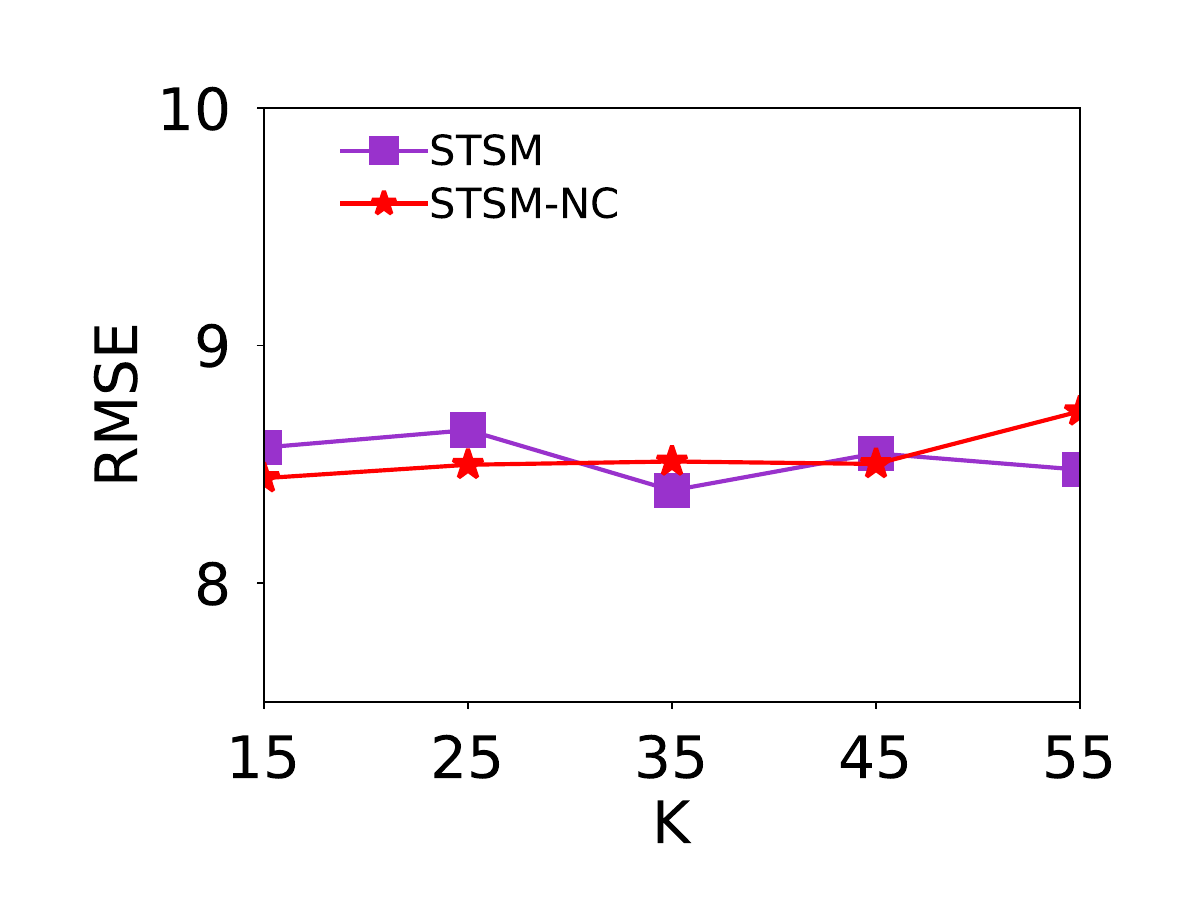}
         \hspace{-2mm}
         \caption{PEMS-07}
         \label{fig:K_07}
     \end{subfigure}
     
     \begin{subfigure}[b]{0.24\textwidth}
         \centering
         \includegraphics[width=\textwidth]{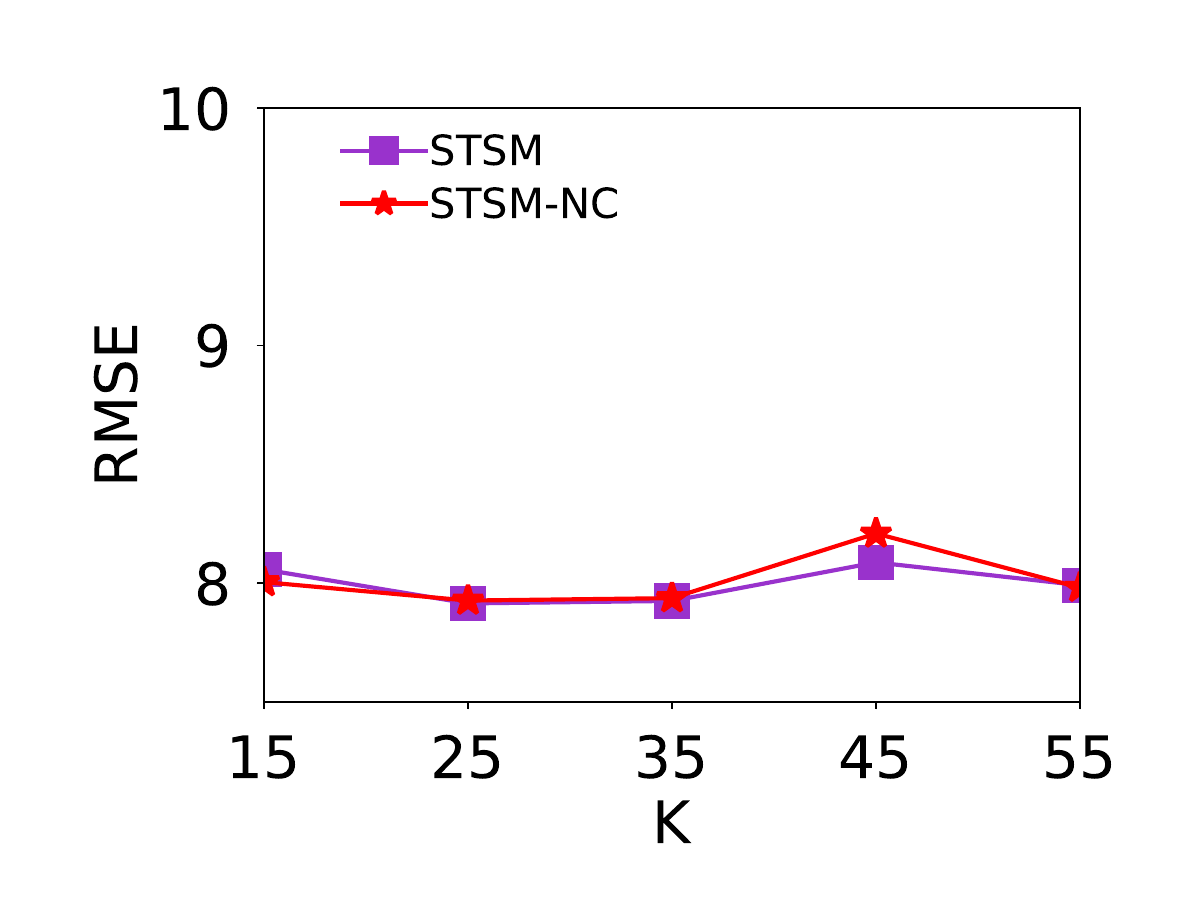}
         \hspace{-2mm}
         \caption{PEMS-08}
         \label{fig:K_08}
     \end{subfigure}
     \hspace{-3mm}
     \begin{subfigure}[b]{0.24\textwidth}
         \centering
         \includegraphics[width=\textwidth]{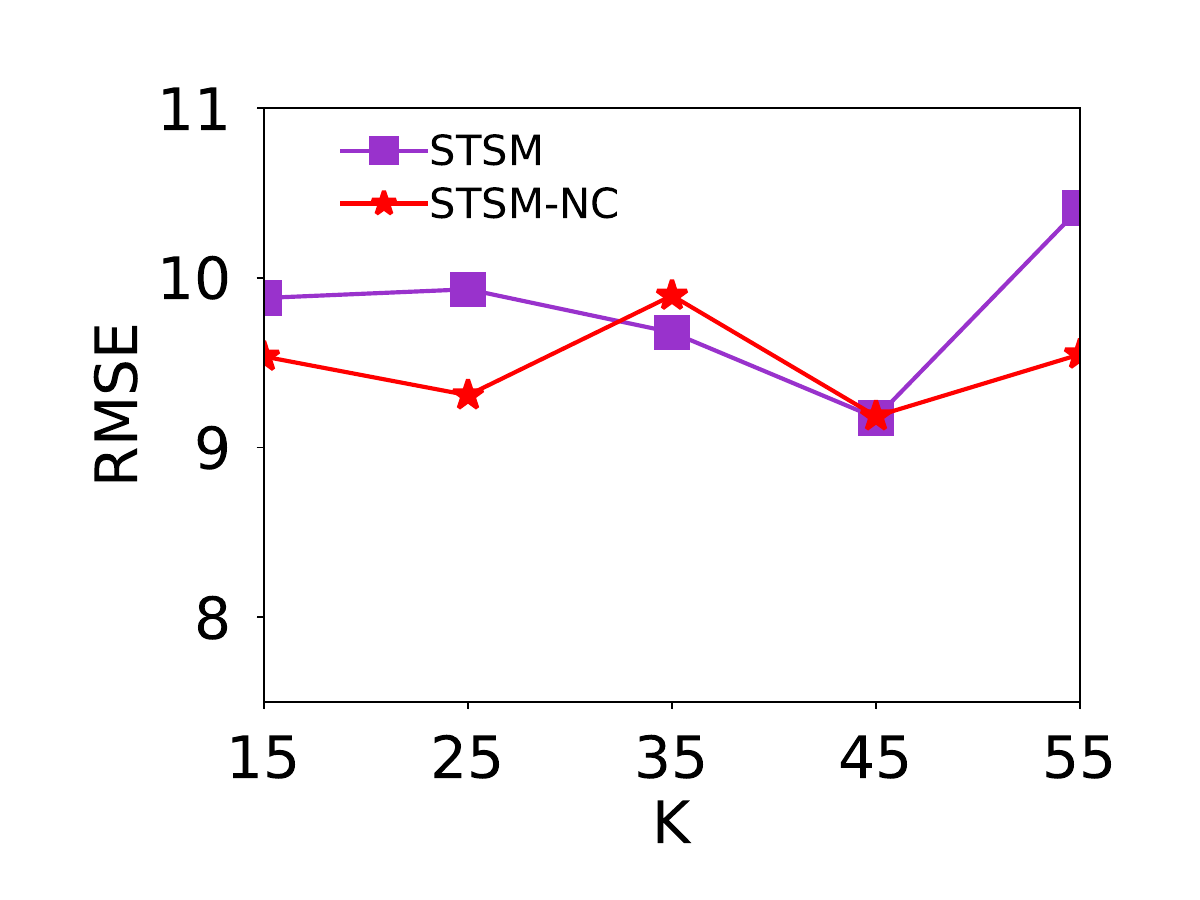}
         \hspace{-2mm}
         \caption{Melbourne}
         \label{fig:K_MEL}
     \end{subfigure}

     \begin{subfigure}[b]{0.24\textwidth}
         \centering
         \includegraphics[width=\textwidth]{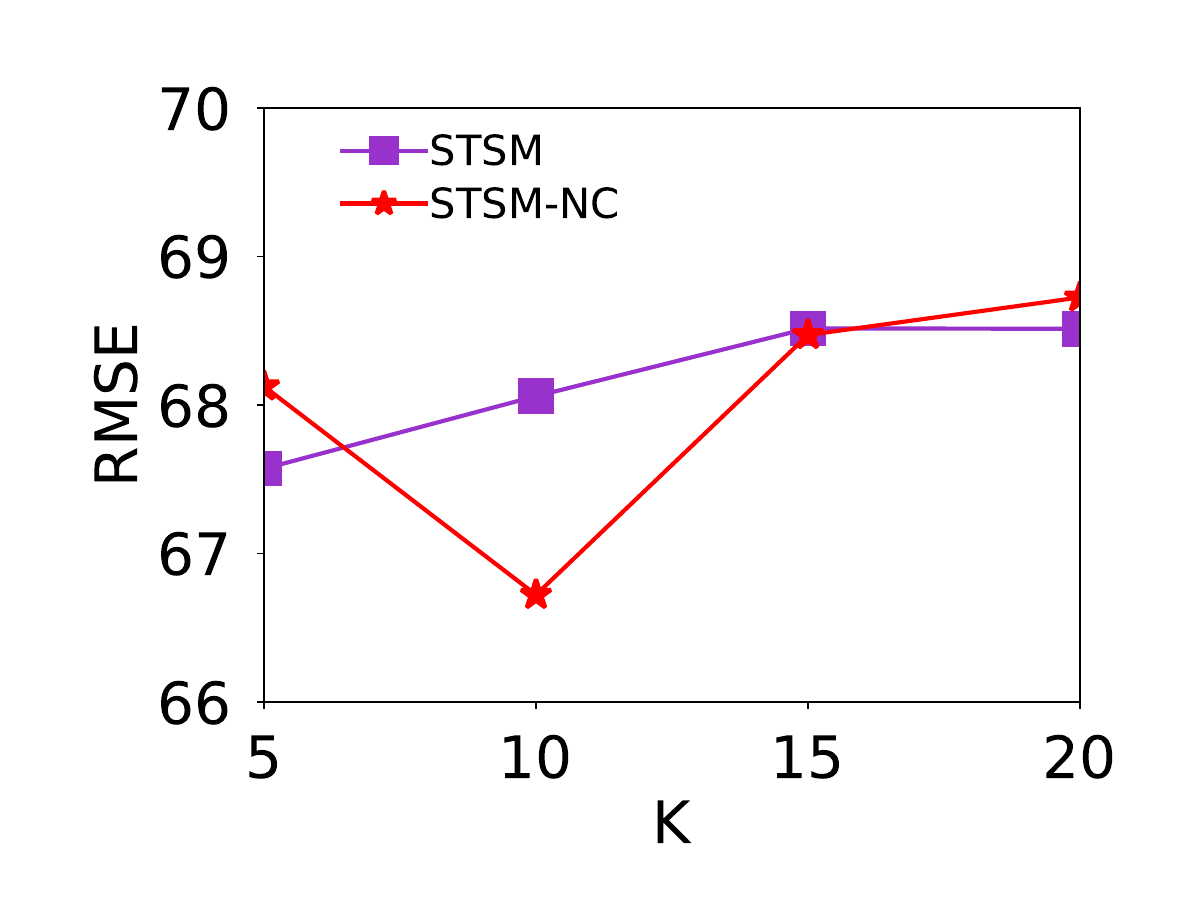}
         \hspace{-2mm}
         \caption{AirQ}
         \label{fig:K_Air}
     \end{subfigure}
        \caption{Model performance vs. K}
        \label{fig:K}
\end{figure}

(1)~\emph{Impact of the number of top similar locations (sub-graphs) $K$.} This parameter determines the number of sub-graphs that may be masked (i.e., impacts the value of $P$ in Eq.~\ref{eq:norm_sim}). It impacts the performance of \model\ and \model-NC because these two model variants use the selective masking module.
The results in Fig.~\ref{fig:K} show that the performance of both \model\ and \model-NC is more stable on freeway traffic datasets than on the other datasets. The reason is that the freeway datasets contain more sensors than the Melbourne and AirQ datasets (i.e., the parameter changing ratio impacts models' sensitivity). 
    
\begin{figure}[h!]
     \centering
     \begin{subfigure}[b]{0.24\textwidth}
         \centering
         \includegraphics[width=\textwidth]{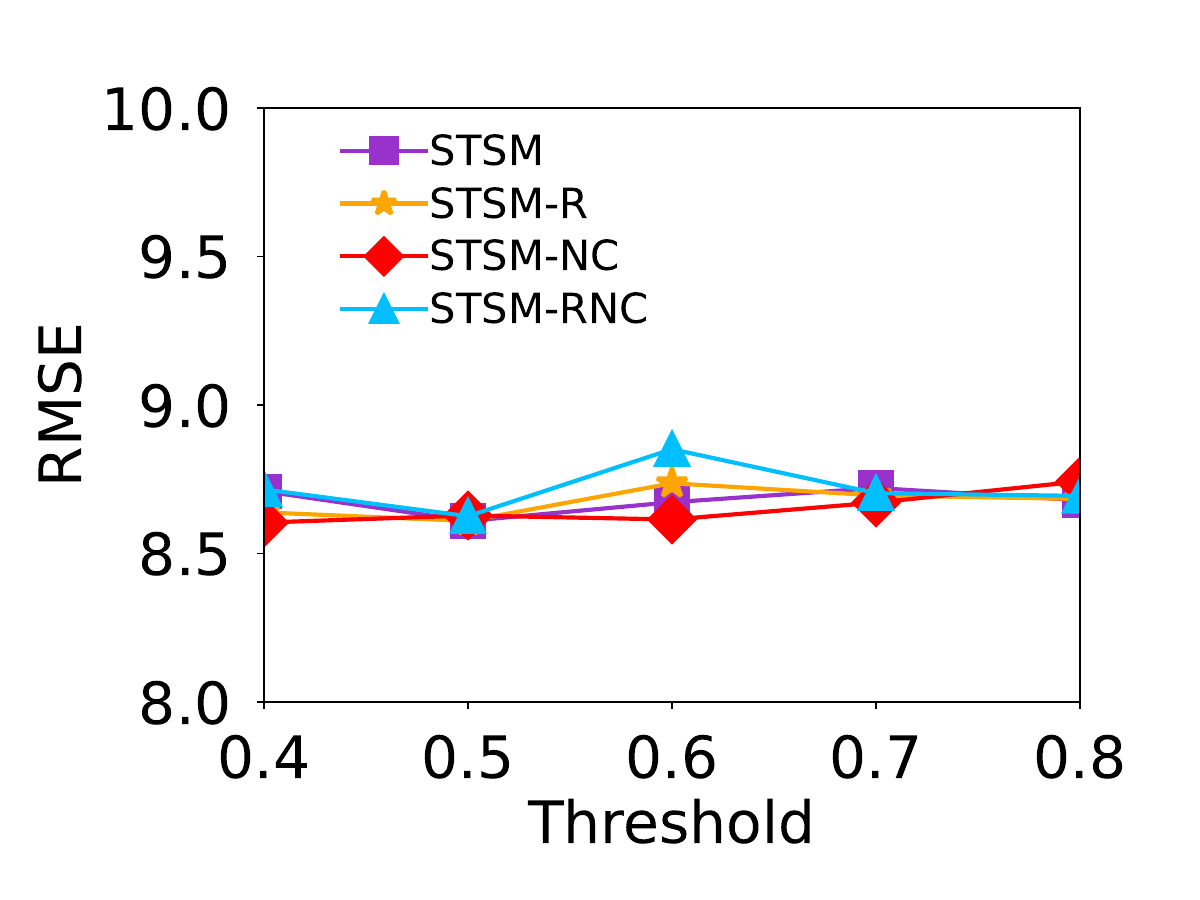}
         \hspace{-2mm}
         \caption{PEMS-Bay}
         \label{fig:adj_ratio_bay}
     \end{subfigure}
     \hspace{-3mm}
     \begin{subfigure}[b]{0.24\textwidth}
         \centering
        \includegraphics[width=\textwidth]{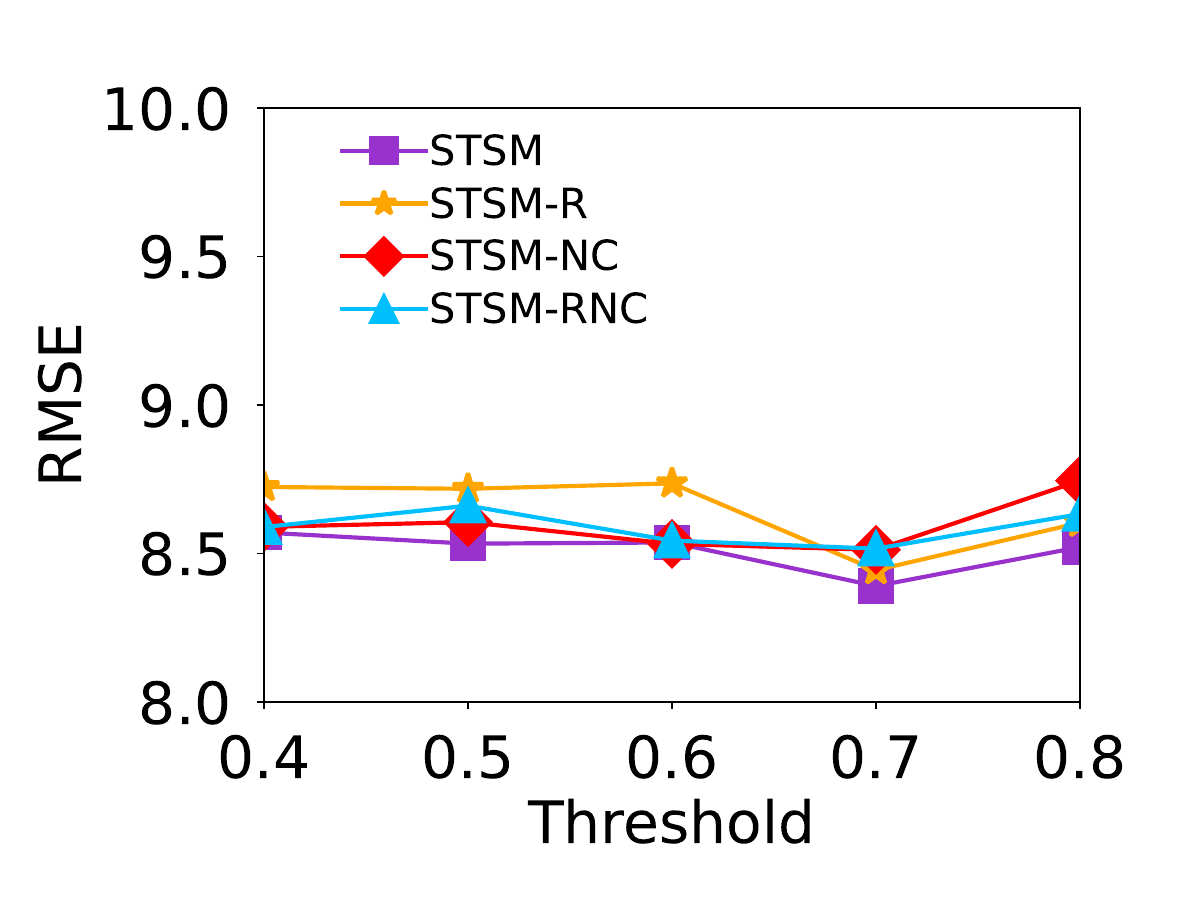}
         \hspace{-2mm}
         \caption{PEMS-07}
         \label{fig:adj_ratio_07}
     \end{subfigure}
     
     \begin{subfigure}[b]{0.24\textwidth}
         \centering
         \includegraphics[width=\textwidth]{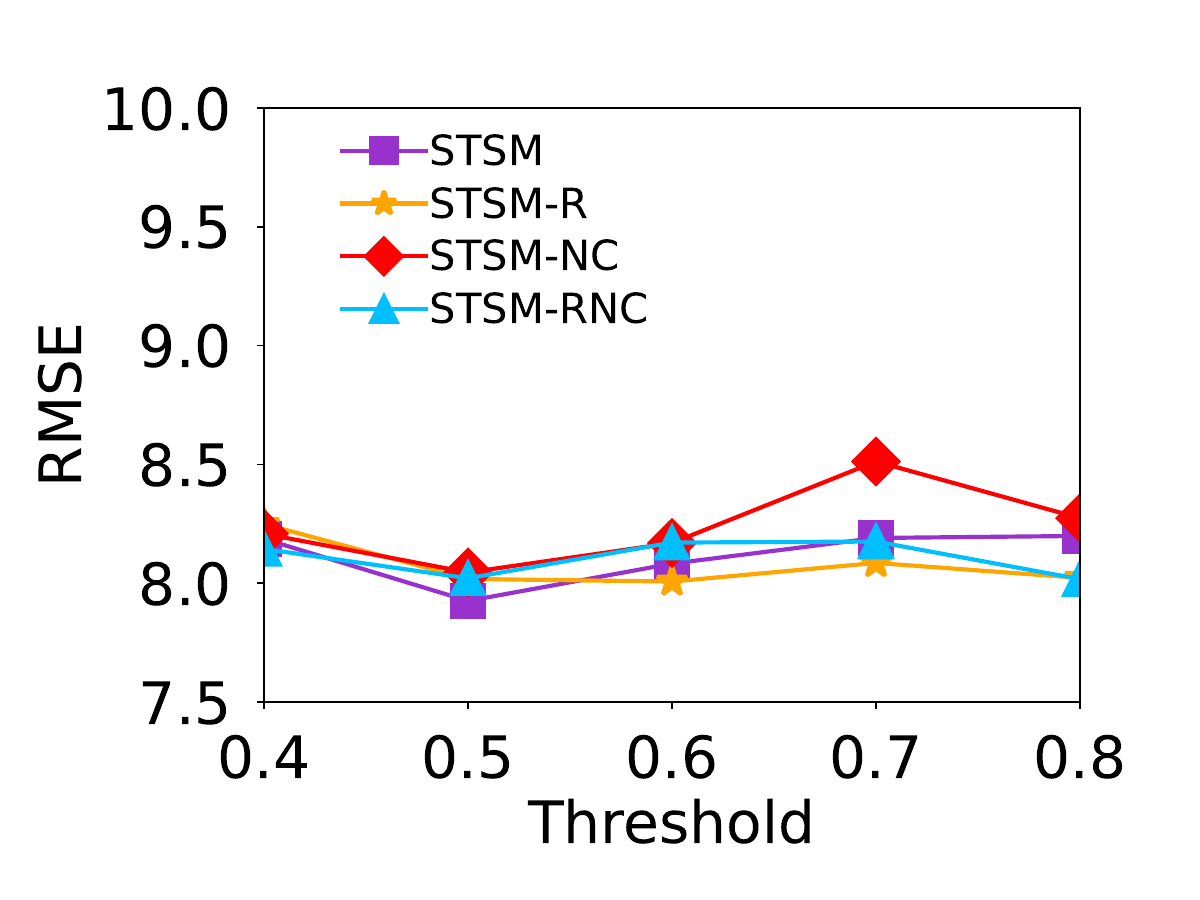}
         \hspace{-2mm}
         \caption{PEMS-08}
         \label{fig:adj_ratio_08}
     \end{subfigure}
     \hspace{-3mm}
     \begin{subfigure}[b]{0.24\textwidth}
         \centering
         \includegraphics[width=\textwidth]{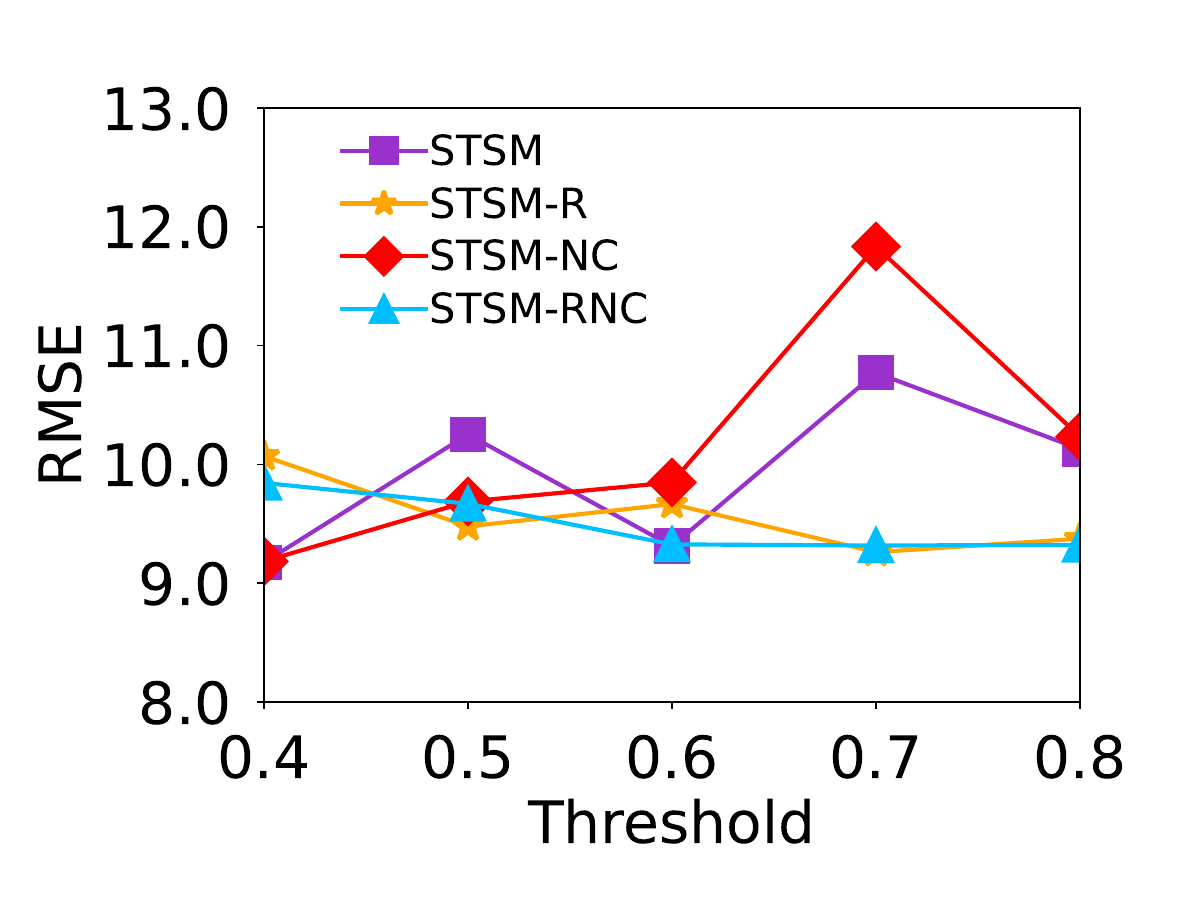}
         \hspace{-2mm}
         \caption{Melbourne}
         \label{fig:adj_ratio_MEL}
     \end{subfigure}

     \begin{subfigure}[b]{0.24\textwidth}
         \centering
         \includegraphics[width=\textwidth]{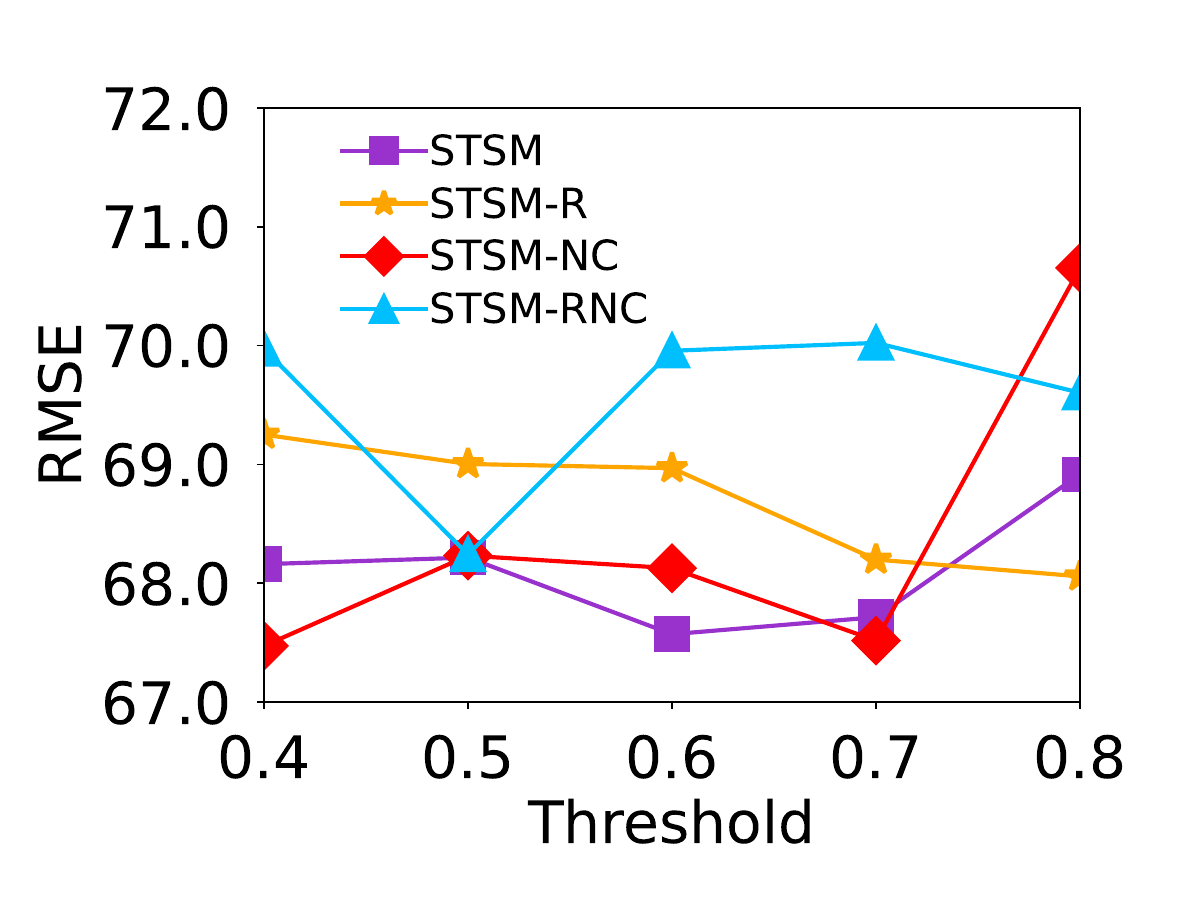}
         \hspace{-2mm}
         \caption{AirQ}
         \label{fig:adj_ratio_Air}
     \end{subfigure}
     \caption{Model performance vs. $\epsilon_{sg}$}
        \label{fig:Adj_ratio}
\end{figure}

(2)~\emph{Impact of the spatial-based matrix's threshold $\epsilon_{sg}$.} This parameter is used to control the size of sub-graphs. When $\epsilon_{sg}$ becomes larger, the sub-graph size becomes smaller (i.e., fewer locations in each sub-graph) because of fewer links in the graph. \model\ and its variants all mask locations based on the sub-graphs, and we test the impact of $\epsilon_{sg}$ on all of them. Fig.~\ref{fig:Adj_ratio} shows the results. \model\ and its variants are again robust about this parameter, especially on the freeway traffic datasets. For Melbourne and AirQ, their smaller numbers of sensors and the complex urban road network information lead to more sensitivities. Note that the fluctuations are quite small compared with the traffic and air pollutant observation values.

\subsubsection{Impact of Space Splits}
\begin{figure}[!h]
     \centering
     \begin{subfigure}[b]{0.25\textwidth}
         \centering
         \includegraphics[width=\textwidth]{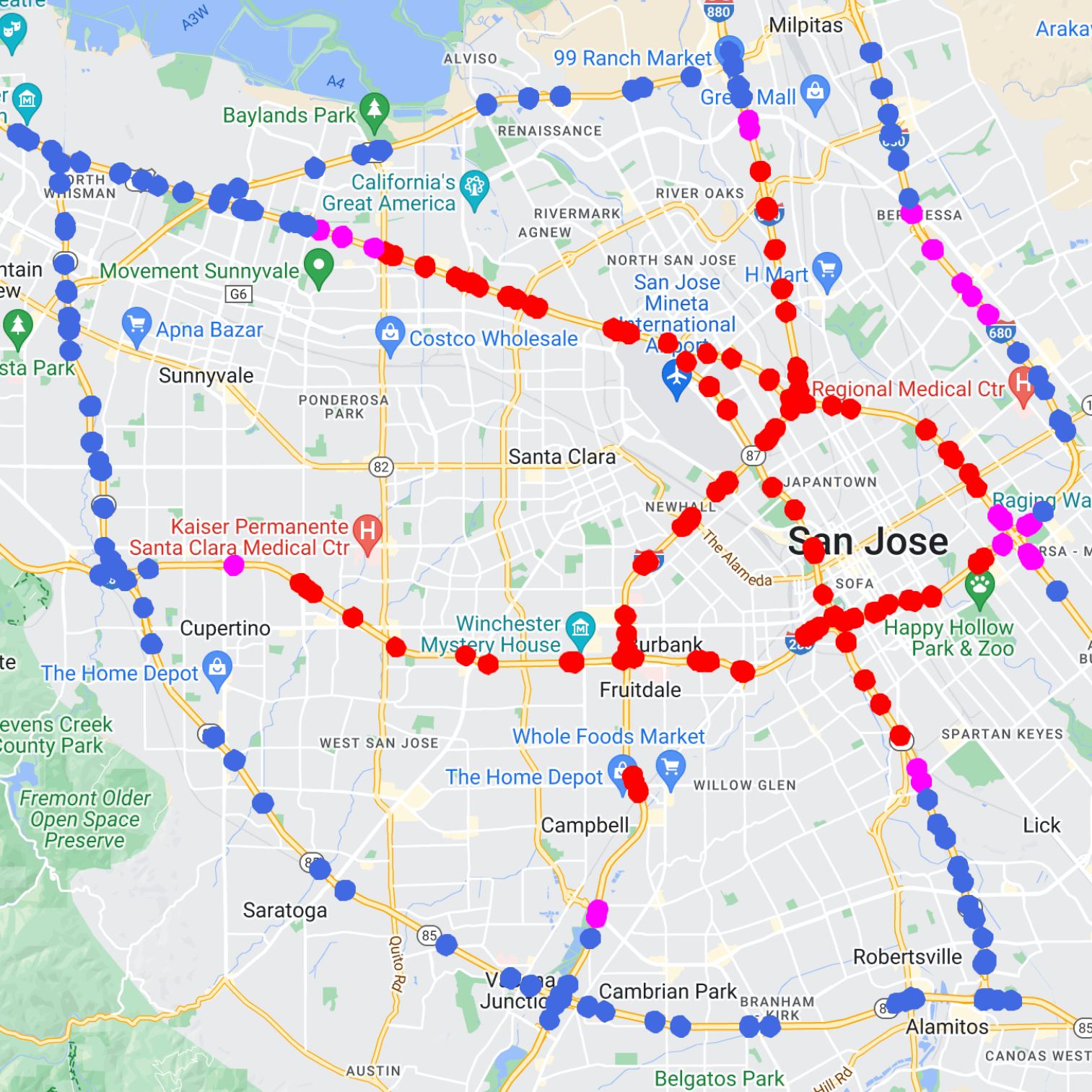}
     \end{subfigure}
     \caption{Visualization of different types of sensor distribution on PEMS-Bay}
     \label{fig:Ring_bay}
\end{figure}

The relative position of the unobserved and observed regions can impact the model result. Our experiments above use horizontal or vertical space splits. To further verify the robustness of \model, considering the circular nature of many city layouts, we study another space splitting strategy (i.e., "ring" splitting) as shown in Fig.~\ref{fig:Ring_bay} - the centre region is the observed region (red dots) for training, the region in the middle ring (pink dots) is for validation, and the outer regions are unobserved (blue dots) for testing. We conduct experiments on PEMS-Bay with this strategy. As shown in Table~\ref{tab:split_5_performance}, \model\ again outperforms all baseline models consistently, with an advantage of up to 9\% in terms of R2.

\begin{table}[h!]
\centering
  \caption{Model performance on PEMS-Bay (Ring Split)}
  \label{tab:split_5_performance}
\begin{tabular}{l|rrrr}
  \hlineB{3}
  Model & RMSE & MAE & MAPE & R2\\
  \hline \hline
  GE-GAN & 32.674 & 26.740 & 0.447 & -10.612\\
  IGNNK & 12.625 & 10.233 & 0.2 & -0.737\\
  INCREASE & 8.558 & 4.807 & 0.125 & 0.190\\
  \model & \textbf{8.462} & \textbf{4.757} & \textbf{0.124} & \textbf{0.208}\\
  \hline
  Improvement & +1.1\% & +1.0\% & +0.8\% & +9.5\%\\
  \hlineB{3}
\end{tabular}
\end{table}

\subsubsection{Impact of Temporal Correlation Learning}
The techniques used to capture temporal correlation patterns can impact the model effectiveness. We have used 1-D CNN in \model\ for its simplicity. In this set of experiments, we further explore the extensibility of \model\ to incorporate advanced temporal correlation modelling techniques. 
We replace 1-D CNN with a transformer encoder (which is an advanced sequence learning model) and a gated fusion module~\cite{gman} to fuse each block's spatial and temporal embeddings. We denote this variant as \emph{\model-trans}. Table~\ref{tab:variant_trans} presents the experimental results on PEMS-Bay. Overall, \model-trans outperforms \model, which verifies the extensibility of \model\ to incorporate advanced correlation pattern learning models.

\begin{table}[h!]
\centering
  \caption{Model performance with advanced temporal correlation learning modules on  PEMS-Bay}
  \label{tab:variant_trans}
\begin{tabular}{l|rrrr}
  \hlineB{3}
  Model & RMSE & MAE & MAPE & R2\\
  \hline \hline
  \model & 8.610 & \textbf{5.237} & 0.130 & 0.231\\
  \model-trans & \textbf{8.562} & 5.251 & \textbf{0.129} & \textbf{0.240}\\
  \hlineB{3}
\end{tabular}
\end{table}

\subsubsection{Impact of Distance Functions}
We used Euclidean distance in our model for efficiency consideration. Road network distance is an alternative choice. To study the impact of the distance function, we compare \model\ (using the Euclidean distance) with two variants: \model-rd-a uses road network distances for computing the adjacency matrices (i.e., $A_s$ and $A_{sg}$) and pseudo-observations, while \model-rd-m uses road network distance for computing the adjacency matrices (i.e., $A_s$ and $A_{sg}$) only.
Table~\ref{tab:distance} shows that \model\ has the best performance among all variants, which verifies that Euclidean distance is efficient and effective for our model. \model-rd-m performs better than \model-rd-a because Euclidean distance leads to better quality of the pseudo-observations.

\begin{table}[h!]
\centering
  \caption{Model performance when using different distance functions on PEMS-Bay}
  \label{tab:distance}
\begin{tabular}{l|rrrr}
  \hlineB{3}
  Model & RMSE & MAE & MAPE & R2\\
  \hline \hline
  \model & \textbf{8.610} & \textbf{5.237} &\textbf{0.130} & \textbf{0.231}\\
  \model-rd-a & 8.993 & 5.426 & 0.134 & 0.158\\
  \model-rd-m & 8.708 & 5.339 & 0.132 & 0.213\\
  \hlineB{3}
\end{tabular}
\end{table}

\section{Conclusions and Future Work}
We proposed a new task - spatial-temporal forecasting for a region of interest without historical observations while this region's adjacent region has such data. We design a novel model named \model\ for the task. We propose a selective masking module based on region, road network and spatial distance features. This module can guide \model\ to mask locations in the adjacent region that have higher similarity with those in the region of interest, which is beneficial for extending the forecasting capability of \model\ to the region of interest. Besides, \model\ exploits contrastive learning to enhance model forecasting efficacy. Extensive experimental results on real-world datasets, including traffic data and air quality data, show that \model\ outperforms adapted state-of-the-art models consistently in forecasting accuracy. This advantage benefits from (1)~the selective masking module which guides the model to mask regions more similar to the region of interest, resulting in better generalization of predictions, (2)~contrastive learning which enhances model accuracy over incomplete data, and (3)~temporal similarity based adjacency matrix computation which strengthens the learning capability of GCNs, allowing messages passing from observed locations to unobserved locations.

We only considered one unobserved region. In the future, we plan to extend \model\ to deal with multiple unobserved regions at the same time.


\begin{acks}
This work is partially supported by Australian Research Council (ARC) Discovery Project DP230101534.
\end{acks}


\bibliographystyle{ACM-Reference-Format}
\bibliography{ref}

\end{document}